\documentclass{article}

 \usepackage[final,nonatbib]{nips_2018}

\usepackage[utf8]{inputenc} 
\usepackage[T1]{fontenc}    
\usepackage{hyperref}       
\usepackage{url}            
\usepackage{booktabs}       
\usepackage{amsfonts}       
\usepackage{nicefrac}       
\usepackage{microtype}      
\usepackage{times}
\usepackage{xcolor}
\usepackage{soul}
\usepackage[small]{caption}
\usepackage[numbers]{natbib}
\usepackage{enumerate}
\usepackage{graphicx}  
\usepackage{amsmath,amssymb}
\usepackage{subcaption}
\usepackage{multirow}
\usepackage{algorithm,algorithmicx,listings} 
\usepackage[noend]{algpseudocode} 

\newcommand{\RR}{I\!\!R} 
\newcommand{\argmax}{\operatornamewithlimits{argmax}}
\newcommand{\argmin}{\operatornamewithlimits{argmin}}
\newcommand{\E}{\mathbb{E}}
\newcommand{\Var}{\mathrm{Var}}
\usepackage[english]{babel}
\usepackage{amsthm}
\newtheorem{theorem}{Theorem}
\newtheorem{lemma}{Lemma}

\title{Assumed Density Filtering Q-learning}

\author{
  Heejin Jeong, Clark Zhang, George J. Pappas\\
  University of Pennsylvania\\
  Philadelphia, PA 19104 \\
  \texttt{\{heejinj, clarkz, pappasg\}@seas.upenn.edu}
  \And
  Daniel D. Lee\\
  Cornell Tech\\
  New York, NY 10044\\
  \texttt{ddl46@cornell.edu}
}
\begin{document}

\maketitle

\begin{abstract}
While off-policy temporal difference (TD) methods have widely been used in reinforcement learning due to their efficiency and simple implementation, their Bayesian counterparts have not been utilized as frequently. One reason is that the non-linear max operation in the Bellman optimality equation makes it difficult to define conjugate distributions over the value functions.
In this paper, we introduce a novel Bayesian approach to off-policy TD methods, called as ADFQ, which updates beliefs on state-action values, Q, through an online Bayesian inference method known as \textit{Assumed Density Filtering}. 
We formulate an efficient closed-form solution for the value update by approximately estimating analytic parameters of the posterior of the Q-beliefs. 
Uncertainty measures in the beliefs not only are used in exploration but also provide a natural regularization for the value update considering all next available actions.
 ADFQ converges to Q-learning as the uncertainty measures of the Q-beliefs decrease and improves common drawbacks of other Bayesian RL algorithms such as computational complexity. 
We extend ADFQ with a neural network. Our empirical results demonstrate that ADFQ outperforms comparable algorithms on various Atari 2600 games, with drastic improvements in highly stochastic domains or domains with a large action space. 
\end{abstract}

\section{Introduction}
Bayesian reinforcement learning is a classic reinforcement learning (RL) technique that utilizes Bayesian inference to integrate new experiences with prior information about the problem in a probabilistic distribution. It explicitly quantifies the uncertainty of the learning parameters unlike standard RL approaches in which uncertainty is unaccounted for. Explicit quantification of the uncertainty can help guide policies that consider the exploration-exploitation trade-off by exploring actions with higher uncertainty more often \citep{Osband2013, Osband2014}. Moreover, it can also regularize posterior updates by properly accounting for uncertainty.

Motivated by these advantages, a number of algorithms have been proposed in both model-based \citep{Dearden2,Strens,Duff,Guez1,Poupart} and model-free Bayesian RL \citep{Dearden1,Engel1,Engel2,Geist,Chowdhary,Ghavamzadeh2006}. 
However, Bayesian approaches to \textit{off-policy temporal difference (TD) learning} have been less studied compared to alternative methods due to difficulty in handling the max non-linearity in the Bellman optimality equation. Previous studies such as Dearden's Bayesian Q-learning \cite{Dearden1} and Kalman Temporal Difference Q-learning (KTD-Q) \cite{Geist} suffer from their computational complexity and scalability. Yet off-policy TD methods such as Q-learning \citep{Watkins} have been widely used in standard RL, including extensions integrating neural network function approximations such as Deep Q-Networks (DQN) \citep{Mnih2013}.

In this paper, we introduce a novel approximate Bayesian Q-learning algorithm, denoted as ADFQ, which updates belief distributions of $Q$ (action-value function) and approximates their posteriors using an online Bayesian inference algorithm known as assumed density filtering (ADF). In order to reduce the computational burden of estimating parameters of the approximated posterior, we propose a method to analytically estimate the parameters. Unlike Q-learning, ADFQ executes a non-greedy update by considering all possible actions for the next state and returns a soft-max behavior and regularization determined by the uncertainty measures of the Q-beliefs. This alleviates overoptimism and instability issues from the greedy update of Q-learning which have been discussed in a number of papers \citep{harutyunyan2016q, tsitsiklis2002convergence,Hasselt, Hasselt_deep}. We prove the convergence of ADFQ to the optimal Q-values by showing that ADFQ becomes identical to Q-learning as all state and action pairs are visited infinitely often. 

ADFQ is computationally efficient and is extended to complex environments with a neural network. There are previous works that implement Bayesian approaches to Deep RL by using uncertainty in the neural network weights and show promising performance in several Atari games \citep{azizzadenesheli2018efficient,o2017uncertainty, Osband2016}. However, these approaches only focus on exploration and uncertainty information does not directly applied to updating RL parameters. Our method differs from these approaches as it explicitly computes the variances of the Q-beliefs and uses them both for exploration and in the value update. Another recent work \citep{Bellemare} proposed a gradient-based categorical DQN algorithm using a distributional perspective. The value distribution in their work represents the inherent randomness of the agent's interactions with its environment. In contrast, the Q-belief defined in ADFQ is a belief distribution of a learning agent on a certain state-action pair. Therefore, only $\epsilon$-greedy is used in their experiments. We evaluate ADFQ with Thompson sampling (TS) \citep{thompson} as well as $\epsilon$-greedy methods in various Atari games and they outperform DQN and Double DQN \citep{Hasselt}.  Particularly, the non-greedy update in ADFQ dramatically improves the performance in domains with a large number of actions and higher stochasticity. Example source code is available online (\hyperlink{https://github.com/coco66/ADFQ}{https://github.com/coco66/ADFQ}).

\section{Background} \label{sec:Background}
\subsection{Assumed Density Filtering}
Assumed density filtering (ADF) is a general technique for approximating the true posterior with a tractable parametric distribution in Bayesian networks. It has been independently rediscovered for a number of applications and is also known as \textit{moment matching}, \textit{online Bayesian learning}, and \textit{weak marginalization} \citep{Opper,Boyen,Maybeck}. Suppose that a hidden variable $\mathbf{x}$ follows a tractable parametric distribution $p(\mathbf{x}|\theta_t)$ where $\theta_t$ is a set of parameters at time $t$. In the Bayesian framework, the distribution can be updated after observing some new data ($D_t$) using Bayes' rule, $\hat{p}(\mathbf{x}|\theta_t,D_t) \propto p(D_t|\mathbf{x},\theta_t)p(\mathbf{x}|\theta_t)$. In online settings, a Bayesian update is typically performed after a new data point is observed, and the updated posterior is then used as a prior for the following iteration.  

When the posterior computed by Bayes' rule does not belong to the original parametric family, it can be approximated by a distribution belonging to the parametric family. In ADF, the posterior is projected onto the closest distribution in the family chosen by minimizing the reverse \textit{Kullback-Leibler} divergence denoted as $KL(\hat{p}||p)$ where $\hat{p}$ is the original posterior distribution and $p$ is a distribution in a parametric family of interest. Thus, for online Bayesian filtering, the parameters for the ADF estimate is given by $\theta_{t+1} = \argmin_{\theta}KL(\hat{p}(\cdot|\theta_t,D_t)||p(\cdot|\theta))$. 

\subsection{Q-learning}
RL problems can be formulated in terms of an MDP described by the tuple, $M=\langle \mathcal{S},\mathcal{A},\mathcal{P}, R,\gamma \rangle$ where $\mathcal{S}$ and $\mathcal{A}$ are the state and action spaces, respectively, $\mathcal{P}:\mathcal{S}\times \mathcal{A}\times\mathcal{S} \rightarrow [0,1]$ is the state transition probability kernel, $R:\mathcal{S}\times\mathcal{A} \rightarrow \RR$ is a reward function, and $\gamma \in[0,1)$ is a discount factor. The value function is defined as $V^{\pi}(s) = \E_{\pi}[\sum^{\infty}_{t=0}\gamma^t R(s_t,a_t)|s_0=s]$ for all $s \in \mathcal{S}$, the expected value of cumulative future rewards starting at a state $s$ and following a policy $\pi$ thereafter. The state-action value ($Q$) function is defined as the value for a state-action pair,  $Q^{\pi}(s,a) = \E_{\pi}[\sum^{\infty}_{t=0}\gamma^t R(s_t,a_t)|s_0=s,a_0 =a]$ for all {\small $s \in \mathcal{S}, a \in \mathcal{A}$}. The objective of a learning agent in RL is to find an optimal policy $\pi^* = \argmax_{\pi}V^{\pi}$. Finding the optimal values, $V^*(\cdot)$ and $Q^*(\cdot,\cdot)$, requires solving the Bellman optimality equation:
\begin{equation}
  Q^*(s,a) = \E_{s'\sim P(\cdot|s,a)}[R(s,a) + \gamma \max_{b\in \mathcal{A}}Q^*(s',b)] \label{eq:Bellman}
\end{equation}
and $V^*(s) = \max_{a \in \mathcal{A}(s)} Q^*(s,a)$ $\forall s \in \mathcal{S}$ where $s'$ is the subsequent state after executing the action $a$ at the state $s$. \textit{Q-learning} is the most popular off-policy TD learning technique due to its relatively easy implementation and guarantee of convergence to an optimal policy \citep{Watkins,Kaelbling}. At time step $t$, Q-learning updates $Q(s_t,a_t)$ after observing a reward $r_t$ and the next state $s_{t+1}$ (one-step TD learning). The update is based on the \textit{TD error} -- a difference between the \textit{TD target}, $r_t+\gamma \max_b Q(s_{t+1},b)$, and the current estimate on $Q(s_t,a_t)$ with a learning rate $\alpha \in (0,1]$:
\begin{equation*}
  Q(s_t,a_t) \leftarrow Q(s_t,a_t) + \alpha\left(r_t+ \gamma \max_{b} Q(s_{t+1},b)-Q(s_t,a_t)\right) 
\end{equation*}
\section{Bayesian Q-learning with Assumed Density Filtering} \label{sec:ADFQ}
\subsection{Belief Updates on Q-values} \label{sec:Qbelief}
We define $Q_{s,a}$ as a Gaussian random variable with mean $\mu_{s,a}$ and variance $\sigma_{s,a}^2$ corresponding to the action value function $Q(s,a)$ for all $s\in \mathcal{S}$ and $a\in \mathcal{A}$. We assume that the random variables for different states and actions are independent and have different means and variances, $ Q_{s,a} \sim \mathcal{N}(\mu_{s,a}, \sigma_{s,a}^2)$ where $ \mu_{s,a} \neq \mu_{s',a'}$ if $s\neq s' \text{ or } a \neq a'$ $\forall s \in \mathcal{S}, \forall a \in \mathcal{A}$.
According to the Bellman optimality equation in Eq.\ref{eq:Bellman}, we can define a random variable for $V(s)$ as $V_s=\max_a Q_{s,a}$. In general, the probability density function for the maximum of Gaussian random variables, $M = \max_{1\leq k \leq N} X_k$ where $X_k \sim \mathcal{N}(\mu_k, \sigma^2_k)$, is no longer Gaussian:
\begin{equation}
  p \left(M = x \right) = \sum^N_{i=1} \frac{1}{\sigma_i} \phi \left( \frac{x-\mu_i}{\sigma_i} \right) \prod^N_{j\neq i} \Phi \left( \frac{x-\mu_j}{\sigma_j} \right) \label{eq:max_distribution}
\end{equation}
where $\phi(\cdot)$ is the standard Gaussian probability density function (PDF) and $\Phi(\cdot)$ is the standard Gaussian cumulative distribution function (CDF) (derivation details are provided in Appendix \ref{app:max_gaussian}). 

For one-step Bayesian TD learning, the beliefs on $\mathbf{Q}= \{Q_{s,a}\}_{\forall s\in \mathcal{S}, \forall a\in \mathcal{A}}$ can be updated at time $t$ after observing a reward $r_t$ and the next state $s_{t+1}$ using Bayes' rule. In order to reduce notation, we drop the dependency on $t$ denoting $s_t = s$, $a_t = a$, $s_{t+1} = s'$, $r_t=r$, yielding the causally related 4-tuple $\mathbf{\tau}=<s,a,r,s'>$. We use the one-step TD target with a small Gaussian white noise, $r+\gamma V_{s'}+W$ where $W \sim \mathcal{N}(0,\sigma^2_w)$, as the likelihood for $Q_{s,a}$. The noise parameter, $\sigma_w$, reflects stochasticity of an MDP. 
We will first derive the belief updates on Q-values with $\sigma_w=0$ for simplicity and then extend the result to the general case. The likelihood distribution can be represented as a distribution over $V_{s'}$ as  $p(r+\gamma V_{s'}|q,\mathbf{\theta}) = p_{V_{s'}}((q-r)/\gamma|s',\mathbf{\theta})$ where $q$ is a value corresponding to $Q_{s,a}$ and $\mathbf{\theta}$ is a set of mean and variance of $\mathbf{Q}$. From the independence assumptions on $\mathbf{Q}$, the posterior update is reduced to an update for the belief on $Q_{s,a}$:
\begin{equation*}
  \hat{p}_{Q_{s,a}}(q|\mathbf{\theta},r,s') \propto  p_{ V_{s'}}\left( \left. \frac{q-r}{\gamma} \right| s',\mathbf{\theta}\right) p_{Q_{s,a}}(q|\mathbf{\theta})
\end{equation*}
Applying Eq.\ref{eq:max_distribution}, the posterior distribution is derived as follows (derivation details in Appendix \ref{app:posterior_derivation}):
\begin{equation}
 \hat{p}_{Q_{s,a}}(q|\mathbf{\theta},r,s') = \frac{1}{Z} \sum_{b\in \mathcal{A}}  \frac{c_{\mathbf{\tau},b}}{\bar{\sigma}_{\mathbf{\tau},b}} \phi \left( \frac{q - \bar{\mu}_{\mathbf{\tau},b}}{\bar{\sigma}_{\mathbf{\tau},b}} \right) \prod_{\substack{b'\in \mathcal{A}\\ b'\neq b}} \Phi \left(\frac{q - (r + \gamma \mu_{s',b'})}{\gamma\sigma_{s',b'}}\right) \label{eq:posterior}
\end{equation}
where $Z$ is a normalization constant and 
\begin{equation}
  c_{\mathbf{\tau},b} = \frac{1}{\sqrt{\sigma_{s,a}^2 + \gamma^2\sigma_{s',b}^2}} \phi \left(\frac{(r+\gamma \mu_{s',b})- \mu_{s,a}}{\sqrt{\sigma_{s,a}^2 + \gamma^2\sigma_{s',b}^2}} \right) \label{eq:weight}
\end{equation}
\begin{equation}
  \bar{\mu}_{\mathbf{\tau},b} = \bar{\sigma}^2_{\mathbf{\tau},b} \left(\frac{\mu_{s,a}}{\sigma_{s,a}^2} + \frac{r+ \gamma \mu_{s',b}}{\gamma^2 \sigma_{s',b}^2} \right)
  \qquad \qquad
  \frac{1}{\bar{\sigma}^2_{\mathbf{\tau},b}} = \frac{1}{\sigma_{s,a}^2} + \frac{1}{\gamma^2\sigma_{s',b}^2} \label{eq:mean_var_bar}
\end{equation}
\begin{figure*}[t!]
  \centering
  \includegraphics[width =\textwidth]{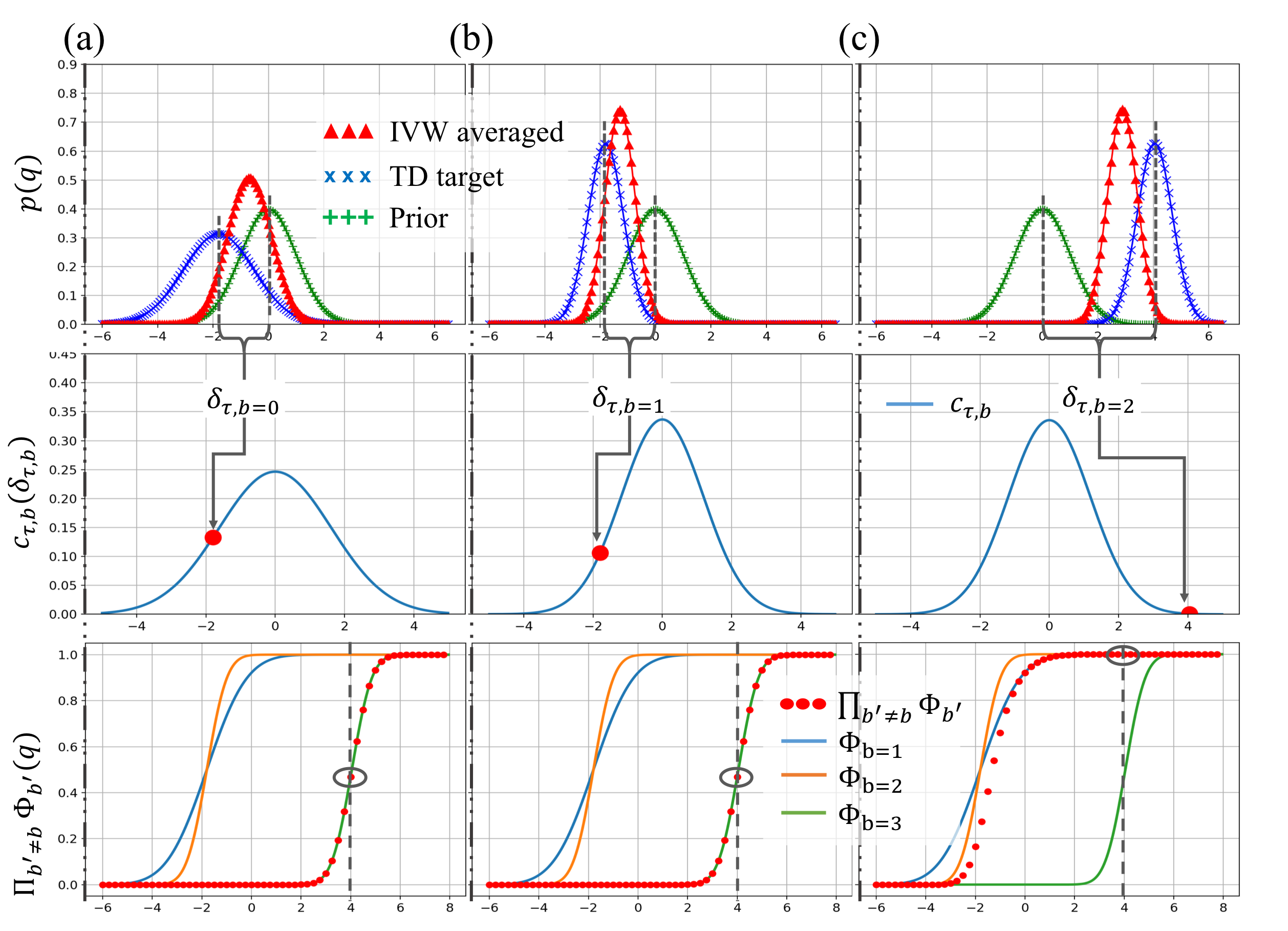}
  \caption{\footnotesize{An example of the belief update in Eq.\ref{eq:posterior} when $|\mathcal{A}|=3, r = 0.0, \gamma=0.9$ and prior (+ green) has $\mu_{s, a} = 0.0, \sigma^2_{s,a}=1.0$. Each column corresponds to a subsequent state and action pair, \textbf{(a) $b=1$}: $\mu_{s',b} = -2.0, \sigma_b^2 = 2.0$, \textbf{(b) $b=2$}: $\mu_{s',b} = -2.0, \sigma_b^2 = 0.5$, \textbf{(c) $b=3$}: $\mu_{s',b} = 4.5, \sigma_b^2 = 0.5$.}}
  \label{fig:illustration}
\end{figure*}
Note that all next actions are considered in Eq.\ref{eq:posterior} unlike the conventional Q-learning update which only considers the subsequent action resulting in the maximum Q-value at the next step ($\max_b Q(s',b)$). This can lead to a more stable update rule as updating with only the maximum Q-value has inherent instability \citep{harutyunyan2016q,tsitsiklis2002convergence}. The Bayesian update considers the scenario where the true maximum Q-value may not be the one with the highest estimated mean, and weights each subsequence Q-value accordingly.
Each term for action $b$ inside the summation in Eq.\ref{eq:posterior} has three important features. First of all, $\bar{\mu}_{\mathbf{\tau},b}$ is an inverse-variance weighted (IVW) average of the prior mean and the TD target mean. Therefore, the Gaussian PDF part becomes closer to the TD target distribution if it has a lower uncertainty than the prior, and vice versa as compared in the first row (a) and (b) of Fig.\ref{fig:illustration}. 
Next, the TD error, $ \delta_{\tau,b} = (r+\gamma \mu_{s',b}) -\mu_{s,a}$, is naturally incorporated in the posterior distribution with the form of a Gaussian PDF in the weight $c_{\tau, b}$. Thus, a subsequent action which results in a smaller TD error contributes more to the update. The sensitivity of a weight value is determined by the prior and target uncertainties. An example case is described in the second row of Fig.\ref{fig:illustration} where $\delta_{\tau,1}=\delta_{\tau,2} > \delta_{\tau,3}$ and $\sigma_{s',1}>\sigma_{s',2}=\sigma_{s',3}$. 
Finally, the product of Gaussian CDFs provides a soft-max operation. The red curve with dots in the third row of Fig.\ref{fig:illustration} represents {\small $\prod_{b'\neq b}\Phi(q|r+ \gamma \mu_{\tau,b'}, \gamma\sigma_{\tau,b'}$)} for each $b$. For a certain $q$ value (x-axis), the term returns a larger value for a larger $\mu_{s',b}$ as seen in the black circles. This result has a similarity with the soft Bellman equation \citep{Ziebart}, but the degree of softness in this case is determined by the uncertainty measures rather than a hyperparameter.

\subsection{Assumed Density Filtering on Q-Belief Updates} 
The posterior distribution in Eq.\ref{eq:posterior}, however, is no longer Gaussian. In order to continue the online Bayesian update, we approximate the posterior with a Gaussian distribution using ADF
. When the parametric family of interest is spherical Gaussian, it is shown that the ADF parameters are obtained by matching moments. Thus, the mean and variance of the approximate posterior are given by those of the true posterior, $\E_{\hat{p}_{Q_{s,a}}}[q]$ and $\Var_{ \hat{p}_{Q_{s,a}}}[q]$, respectively. It is fairly easy to derive the mean and variance when $|\mathcal{A}|=2$. The derivation is presented in Appendix \ref{app:mgf}. However, to our knowledge, there is no analytically tractable solution for $|\mathcal{A}|>2$.

When $\sigma_w > 0$, the expected likelihood is obtained by solving $\int_{\mathbb{R}} p(r+\gamma V_{s'}+w|q,\theta)p_W(w) dw$ which is an integral of a similar form with the posterior in Eq.\ref{eq:posterior}. Therefore, a closed-form expression is also not available in general except when $|\mathcal{A}|=2$ (see Appendix \ref{app:stochastic_mdp}).

In the next sections, we prove the convergence of the means to the optimal Q-values for the case $|\mathcal{A}|=2$ with the exact solutions for the ADF parameters. Then, we show how to derive an analytic approximation for the ADF parameters which becomes exact in the small variance limit.

\subsection{Convergence to Optimal Q-values}
The convergence theorem of the Q-learning algorithm has previously been proven \citep{Watkins}. We, therefore, show that the online Bayesian update using ADF with the posterior in Eq.\ref{eq:posterior} converges to Q-learning when $|\mathcal{A}|=2$. We apply an approximation from Lemma \ref{lemma:normal_approx} in order to prove Theorem \ref{theorem:convergence_numeric}. Proofs for Lemma \ref{lemma:normal_approx} and Theorem \ref{theorem:convergence_numeric} are presented in Appendix \ref{app:proofs}.
\begin{lemma} \label{lemma:normal_approx}
Let $X$ be a random variable following a normal distribution, $\mathcal{N}(\mu,\sigma^2)$. Then we have:
\begin{equation}
	\lim_{\sigma \rightarrow 0 } \left[\Phi \left(\frac{x-\mu}{\sigma}\right) - \exp\left\{ -\frac{1}{2}\left[ -\frac{x-\mu}{\sigma} \right]_+^2 \right\} \right] = 0
\end{equation}
where $[x]_+=\max(0,x)$ is the {\em ReLU} nonlinearity.
\end{lemma}
\begin{theorem} \label{theorem:convergence_numeric}
Suppose that the mean and variance of $Q_{s,a}$ $\forall s \in \mathcal{S}, \forall a \in \mathcal{A}$ are iteratively updated by the mean and variance of $\hat{p}_{Q_{s,a}}$ after observing $r$ and $s'$ at every step. When $|\mathcal{A}|=2$, the update rule of the means is equivalent to the Q-learning update if all state-action pairs are visited infinitely often and the variances approach 0.
In other words, at the $k$th update on $\mu_{s,a}$:
\[
	\lim_{k\rightarrow \infty, \{\sigma\} \rightarrow 0}\mu_{s,a;k+1} = \left(1-\alpha_{\tau;k} \right) \mu_{s,a;k} + \alpha_{\tau;k} \big(r+\gamma \max_{b\in \mathcal{A}} \mu_{s',b;k}\big)
   \]
 where $\alpha_{\tau;k}=\sigma_{s,a;k}^2 / \left( \sigma_{s,a;k}^2 + \gamma^2 \sigma_{s',b^+;k}^2 + \sigma_w^2 \right)$ and $b^+ = \argmax_{b\in\mathcal{A}} \mu_{s',b}$.
\end{theorem}
Interestingly, $\alpha_{\tau}$ approaches 1 when $\sigma_{s,a}/\sigma_{s',b^+}\rightarrow \infty$ and 0 when $\sigma_{s,a}/\sigma_{s',b^+}\rightarrow 0$ for $\sigma_w=0$. Such behavior remains when $\sigma_w >0$ but $\alpha_{\tau}$ eventually approaches 0 as the number of visits to $(s,a)$ goes to infinity.
This not only satisfies the convergence condition of Q-learning but also provides a natural learning rate -- the smaller the variance of the TD target (the higher the confidence), the more $Q_{s,a}$ is updated from the target information rather than the current belief. We show empirical evidence that the contraction condition on variance in Theorem \ref{theorem:convergence_numeric} holds in Appendix \ref{app:mgf}.

\section{Analytic ADF Parameter Estimates} \label{sec:small_variance}
When $|\mathcal{A}|>2$, the update can be solved by numerical approximation of the true posterior mean and variance using a number of samples. However, its computation becomes unwieldy due to the large number of samples needed for accurate estimates. This becomes especially problematic with small variances as the number of visits to corresponding state-action pairs grows. In this section, we show how to accurately estimate the ADF parameters using an analytic approximation.  This estimate becomes exact in the small variance limit.

\subsection{Analytic Approximation of Posterior}
Applying Lemma \ref{lemma:normal_approx} to the Gaussian CDF terms in Eq.\ref{eq:posterior}, the posterior is approximated to the following:
\begin{equation}
\tilde{p}_{Q_{s,a}}(q) = \frac{1}{Z}\sum_{b\in \mathcal{A}}\frac{ c_{\tau, b}}{\sqrt{2\pi}\bar{\sigma}_{\tau,b}}\exp\left\{ -\frac{ \left(q-\bar{\mu}_{\tau, b} \right)^2}{2\bar{\sigma}_{\tau,b}^2} - \sum_{b' \neq b} \frac{\left[r+\gamma \mu_{s',b'}-q \right]_+^2}{2\gamma^2\sigma^2_{s',b'}}\right\}  \label{eq:posterior_approxV1}
\end{equation}
Similar to Laplace's method, we approximate each term as a Gaussian distribution by matching the maximum values as well as the curvature at the peak of the distribution. In other words, the maximum of the distribution is modeled locally near its peak by the quadratic concave function:
\begin{equation}
	 -\frac{ \left(q-\bar{\mu}_{\tau, b} \right)^2}{2\bar{\sigma}_{\tau,b}^2} - \sum_{b' \neq b} \frac{\left[r+\gamma \mu_{s',b}-q \right]_+^2}{2\gamma^2\sigma^2_{s',b}} \approx -\frac{(q-\mu_{\tau,b}^*)^2}{2{\sigma^*_b}^2} \label{eq:Ab}
\end{equation}
We find $\mu^*_{\tau,b}$ and $\sigma^*_{\tau,b}$ by matching the first and the second derivatives, respectively (the coefficient of the quadratic term gives the local curvature):
\begin{equation}
\frac{\mu^*_{\tau,b} - \bar{\mu}_{\tau,b}}{\bar{\sigma}^2_{\tau,b}} = \sum_{b'\neq b} \frac{\left[ r + \gamma \mu_{s', b'}- \mu_{\tau,b}^*\right]_+}{\gamma^2 \sigma_{s',b'}^2} \qquad  
\frac{1}{{\sigma_{\tau,b}^*}^2} =  \frac{1}{\bar{\sigma}^2_{\tau, b}} + \sum_{b'\neq b} \frac{H\left(r + \gamma \mu_{s',b'} - \mu^*_{\tau,b}\right)}{\gamma^2\sigma^2_{s',b'}} \label{eq:peak}
\end{equation}
where $H(\cdot)$ is a Heaviside step function. The self-consistent piece-wise linear equation for $\mu^*_{\tau,b}$ can be rewritten as follows:
\begin{equation*}
\mu_{\tau,b}^* = \left(\frac{1}{\bar{\sigma}_{\tau,b}^2}+\sum_{b'\neq b}\frac{H(r+\gamma\mu_{s',b'}-\mu^*_{\tau,b})}{\gamma^2\sigma^2_{s',b'}} \right)^{-1} \left(\frac{\bar{\mu}_{\tau,b}}{\bar{\sigma}_{\tau,b}^2}+\sum_{b'\neq b}\frac{(r+\gamma\mu_{s',b'})}{\gamma^2\sigma^2_{s',b'}}H(r+\gamma\mu_{s',b'}-\mu^*_{\tau,b}) \right) \label{eq:peak_mean}
\end{equation*}
 This is an IVW average mean of the prior, the TD target distribution of $b$, and other TD target distributions whose means are larger than $\mu^*_{\tau,b}$. The height of the peak is computed for $q=\mu_{\tau,b}^*$,
\begin{equation}
 k^*_{\tau,b} = \frac{c_{\tau, b}\sigma^*_{\tau,b}}{\bar{\sigma}_{\tau,b}}\exp \left\{-\frac{ \left(\mu_{\tau,b}^*-\bar{\mu}_{\tau, b} \right)^2}{2\bar{\sigma}_{\tau,b}^2} - \sum_{b' \neq b} \frac{\left[r+\gamma \mu_{s',b'}-\mu_{\tau,b}^* \right]_+^2}{2\gamma^2\sigma^2_{s',b'}} \right\} \label{eq:magnitude}
\end{equation}
The final approximated distribution is a Gaussian mixture model with $\mu^*_{\tau,b}, \sigma^*_{\tau,b}, w^*_{\tau,b}$ as mean, variance, and weight, respectively, for all $b \in \mathcal{A}$ where $w^*_{\tau,b}=k^*_{\tau,b}/\sum_{b'}k^*_{\tau,b'}$. 
Therefore, we update the belief distribution over $Q_{s,a}$ with the mean and variance of the Gaussian mixture model:
\begin{equation}
  \E_{\tilde{p}}[q] = \sum_{b\in \mathcal{A}} w_{\tau,b}^* \mu^*_{\tau,b} \quad \qquad
 \Var_{\tilde{p}}[q]  = \sum_{b\in \mathcal{A}} w_{\tau,b}^*{\sigma_{\tau,b}^*}^2 + \sum_{b\in \mathcal{A}} w_{\tau,b}^* {\mu^*_{\tau,b} }^2  - \left( \E_{\tilde{p}}[q]\right)^2 \label{eq:approx_v1}
\end{equation}
The final mean is the weighted sum of each individual mean with a weight from $k^*_{\tau,b}$ and the final variance is the weighted sum of each individual variance added to a non-negative term accounting for the dispersion of the means. 
As shown in Eq.\ref{eq:magnitude}, the weights are determined by TD errors, variances, relative distances to larger TD targets. It has the TD error penalizing term, $c_{\tau,b}$, and also decreases as the number of TD targets larger than $\mu^*_{\tau,b}$ increases. Therefore, the weight provides a softened maximum property over $b$.
\begin{algorithm}[t!]
\caption{ADFQ algorithm}
\begin{algorithmic}[1]
\small
\State Initialize randomly $\mu_{s,a}$, $\sigma_{s,a}$ $\forall s \in \mathcal{S}$ and $\forall a \in \mathcal{A}$
\For{each episode}
\State Initialize $s_0$
\For{each time step $t$}
\State Choose an action, $a_t \sim$ $\pi^{action}(s_t;\theta_t)$ 
\State Perform the action and observe $r_t$ and $s_{t+1}$
\For{each $b \in \mathcal{A}$}
\State Compute $\mu^*_{\tau,b}$, $\sigma^*_{\tau,b}$, $k^*_{\tau,b}$ using Eq.\ref{eq:peak}-\ref{eq:magnitude}
\EndFor
\State Update $\mu_{s_t, a_t}$ and $\sigma_{s_t,a_t}$ using Eq.\ref{eq:approx_v1}
\EndFor
\EndFor
\end{algorithmic} \label{table:adfq}
\end{algorithm}
The final algorithm is summarized in Table.\ref{table:adfq}. Its space complexity is $O(|\mathcal{S}||\mathcal{A}|)$. The computational complexity of each update is $O(|\mathcal{A}|^2)$ 
which is higher than Q-learning but only by a factor of $|\mathcal{A}|$ and constant in the number of states. 

\subsection{Approximate Likelihood }
In an asymptotic limit of $\sigma_w/\sigma_{s',b} \rightarrow 0$, $\forall b\in \mathcal{A}$ and $|\mathcal{A}|=2$, the expected likelihood distribution for $\sigma_w >0$ is similar to $p(r+\gamma V_{s'}|q, \theta)$ but the variance of its Gaussian PDF term is $\gamma^2\sigma_{s',b}^2 + \sigma_w^2$ instead of $\gamma^2\sigma_{s',b}^2$ (see Appendix \ref{app:stochastic_mdp} for details):
\begin{align}
\sum_{b\in \mathcal{A}} \frac{\gamma}{ \sqrt{\gamma^2\sigma_{s',b}^2+\sigma^2_w}} \phi \left( \frac{q - (r + \gamma \mu_{s',b})}{\sqrt{\gamma^2\sigma_{s',b}^2+\sigma^2_w}} \right) \prod_{b'\neq b, b'\in \mathcal{A}} \Phi \left(\frac{q - (r + \gamma \mu_{s',b'})}{\gamma\sigma_{s',b'}}\right) \label{eq:ll_sto}
\end{align}
Extending this result to the general case ($|\mathcal{A}|=n$ for $n \in \mathbb{N}$), the posterior distribution, $\hat{p}_{Q_{s,a}}(q)$, for $\sigma_w >0$ is same with Eq.\ref{eq:posterior} but $\gamma^2\sigma_{s',b}^2$ is replaced by $\gamma^2\sigma_{s',b}^2 + \sigma_w^2$ in $c_{\tau,b}$, $\bar{\mu}_{\tau,b}$, and $\bar{\sigma}_{\tau,b}$ (Eq.\ref{eq:weight}-\ref{eq:mean_var_bar}). Therefore, $\mu_{\tau,b}^*$, $\sigma_{\tau,b}^*$, and $k_{\tau,b}^*$ in the ADFQ algorithm (Table.\ref{table:adfq}) are also changed accordingly.

\subsection{Convergence of ADFQ}
Theorem \ref{theorem:convergence_numeric} extends to the ADFQ algorithm (Proof in Appendix \ref{app:proofs}). The contraction behavior of the variances in the case of Theorem \ref{theorem:convergence_numeric} is also empirically observed in ADFQ.
\begin{theorem} \label{theorem:convergence_adfq}
The ADFQ update on the mean $\mu_{s,a}$ $\forall s\in \mathcal{S}$, $\forall a \in \mathcal{A}$ for $|\mathcal{A}|=2$ is equivalent to the Q-learning update if the variances approach 0 and if all state-action pairs are visited infinitely often. In other words, we have :
\[
	\lim_{k\rightarrow \infty, \{\sigma\} \rightarrow 0}\mu_{s,a;k+1} = \left(1-\alpha_{\tau;k} \right) \mu_{s,a;k} + \alpha_{\tau;k} \left(r+\gamma \max_{b\in \mathcal{A}} \mu_{s',b;k}\right)
   \]
 where $\alpha_{\tau;k}=\sigma_{s,a;k}^2 / \left( \sigma_{s,a;k}^2 + \gamma^2 \sigma_{s',b^+;k}^2 + \sigma_w^2\right)$ and $b^+ = \argmax_{b\in\mathcal{A}} \mu_{s',b}$.
\end{theorem}
As we have observed the behavior of $\alpha_{\tau}$ in Theorem \ref{theorem:convergence_numeric}, the learning rate $\alpha_{\tau}$ again provides a natural learning rate with the ADFQ update. We can therefore think of Q-learning as a special case of ADFQ.

\begin{figure}[t!]
\centering
\includegraphics[width=\columnwidth]{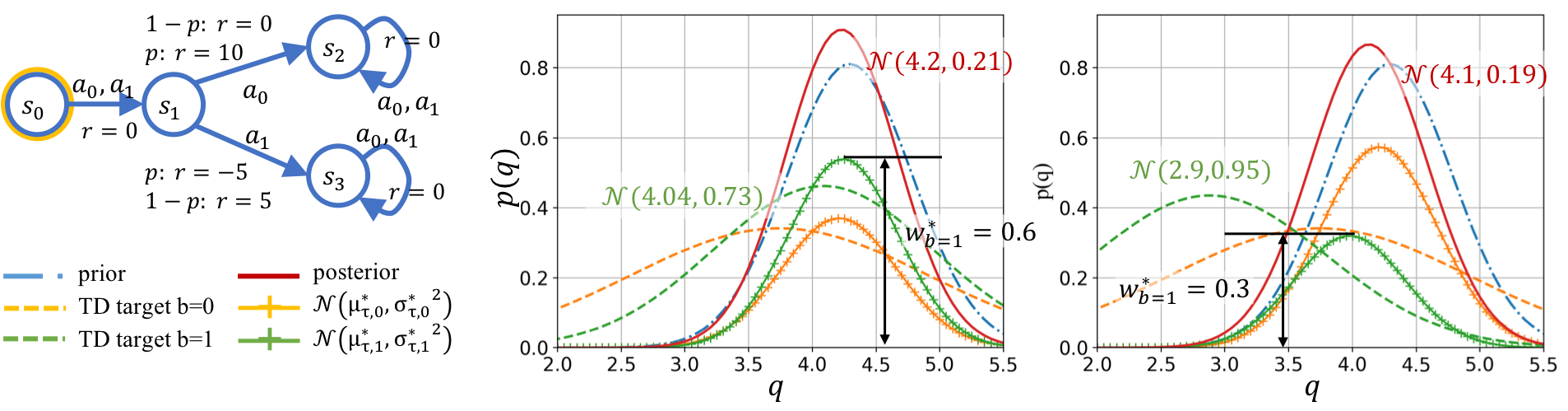}
\caption{\footnotesize A simple MDP with stochastic rewards and ADFQ update example for $s_{t+1}=s_0, a_{t+1}=a_0$, (left) $r_t=5$, (right) $r_t=-5$}
\label{fig:simple_mdp}
\end{figure}
\section{Demonstration in Discrete MDPs} \label{sec:demo}
To demonstrate the behavior of the ADFQ update, we look at the simple MDP ($\gamma=0.9$) in Fig.\ref{fig:simple_mdp} at a specific iteration. An episode starts at $s_0$ and terminates at either $s_2$ or $s_3$. At $s_1$, each action returns a stochastic reward with $p=0.2$. The optimal deterministic policy at $s_1$ is $a_1$. 
Suppose an RL learner has already visited $(s_1, a_1)$ 3 times and obtained a reward of $r=5$ every time. Now it is on the $t$-th iteration with $(s_1, a_1)$
The plots in Fig.\ref{fig:simple_mdp} show the ADFQ update for $Q_{s_0,a_0}$ at $t+1$ when $r_t=+5$ (left) and $r_t=-5$ (right).
When it receives a less expected reward, $-5$, at $t$, $\sigma_{s_1,a_1;t}$ is updated to a larger value than the one in the $r_t=+5$ case. Then, the episode is terminated and the next episode starts at $s_{t+1}=s_0$, $a_{t+1}=a_0$. ADFQ considers both $Q_{s_1, a_0}$ and $Q_{s_1,a_1}$ for updating $Q_{s_0,a_0}$. Due to the relatively large TD error and variance of $Q_{s_1,a_1}$, a lower value is assigned to $w^*_{\tau,b=1}$. In this same scenario, Q-learning would update $Q(s_0, a_0)$ only from $Q(s_1,a_0)$ and regulate the update amount with the learning rate which is usually fixed or determined by the number of visits.

\begin{figure}[b!]
\centering
\includegraphics[height=4cm]{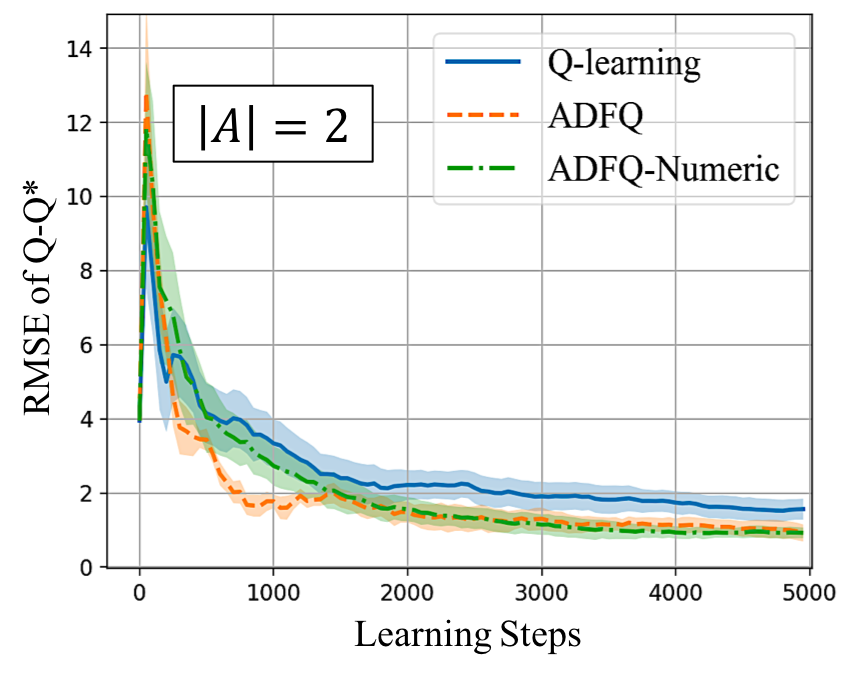}
\hspace{1cm}
\includegraphics[height=4cm]{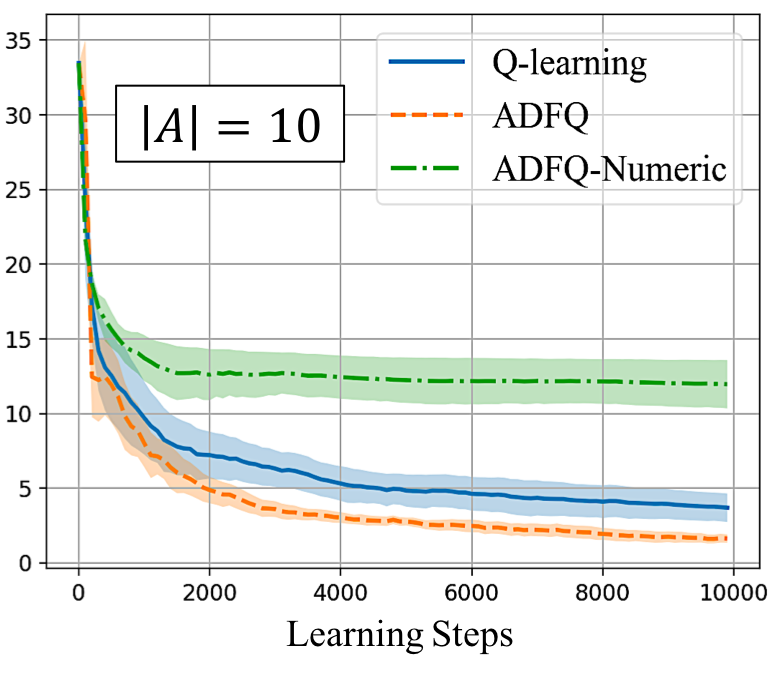}
\caption{\footnotesize Convergence to $Q^*$ in an MDP $|\mathcal{A}|=2$ (left) and an MDP $|\mathcal{A}|=10$ (right)}
\label{fig:rmse}
\end{figure}
In order to show the benefits of the update rule, we examined Q-learning, ADFQ, and a numerical approximation of the mean and variance of Eq.\ref{eq:posterior} (denoted as ADFQ-Numeric) for the convergence to the optimal Q-values in the presented MDP and a similar MDP but with 10 terminating states and 10 actions. Random exploration is used in order to evaluate only the update part of each algorithm. During learning, we computed the root mean square error (RMSE) between the estimated Q-values (or means) and $Q^*$, and plotted the averaged results over 5 trials in Fig.\ref{fig:rmse}. As shown, ADFQ converged to the optimal Q-values quicker than Q-learning in both cases and showed more stable performance. ADFQ-Numeric suffers from correctly estimating the parameters when its variances become small as it is previously pointed out, and resulted a poor convergence result in the large MDP. 

\begin{figure}[tb!]
\centering
\includegraphics[width=0.75\columnwidth]{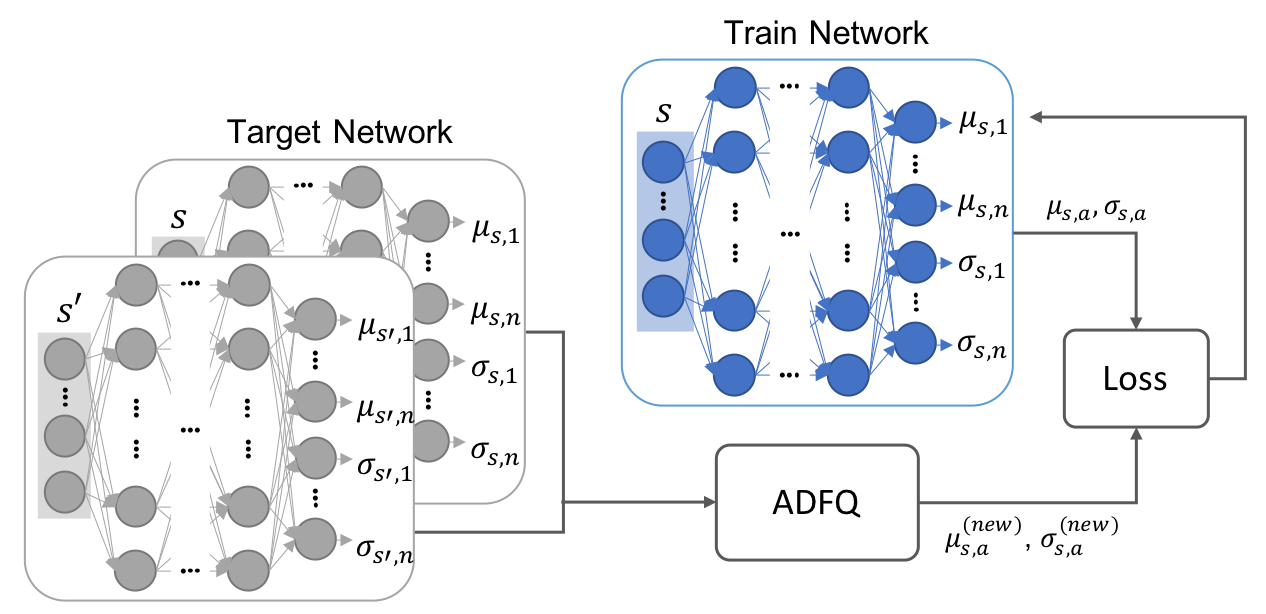}
\caption{\footnotesize A neural network model for ADFQ}
\label{fig:adfqn}
\end{figure}
\section{ADFQ with Neural Networks}
In this section, we extend our algorithm to complex environments with neural networks similar to Deep Q-Networks (DQN) proposed in \citep{Mnih2013}. In the Deep ADFQ model with network parameters $\xi$, the output of the network is mean $\mu(s,a;\xi)$ and variance $\sigma^2(s,a;\xi)$ of each action for a given state $s$ as shown in Fig.\ref{fig:adfqn}. In practice, we use $-\log(\sigma_{s,a})$ instead of $\sigma^2_{s,a}$ for the output to ensure positive values for the variance. As in DQN, we have a train network ($\xi$) and a target network ($\xi'$). Mean and variance for $s$ and $s'$ from the target network are used as inputs into the ADFQ algorithm to compute the desired mean, $\mu^{ADFQ}$, and standard deviation, $\sigma^{ADFQ}$ for the train network. We used prioritized experience replay \citep{schaul2015prioritized} and a combined Huber loss functions of mean and variance.

In order to demonstrate the effectiveness of our algorithm, we tested on six Atari games, Enduro ($|\mathcal{A}|=9$), Boxing ($|\mathcal{A}|=18$), Pong ($|\mathcal{A}|=6$), Asterix ($|\mathcal{A}|=9$), Kung-Fu Master ($|\mathcal{A}|=14$), and Breakout ($|\mathcal{A}|=4$), from the OpenAI gym simulator \citep{openaigym}. For baselines, we used DQN and Double DQN with prioritized experience replay implemented in OpenAI baselines \citep{baselines} with their default hyperparameters for all games. We used $\epsilon$-greedy action policy with $\epsilon$ annealed from 1.0 to 0.01 for the baselines as well as ADFQ. In ADFQ, the greedy selection is performed on the mean values instead of Q-values. Additionally, we examined  \textit{Thompsing Sampling} (TS) \cite{thompson} for ADFQ which selects $a_t = \argmax_a q_{s_t,a}$ where $q_{s_t,a} \sim p_{Q_{s_t,a}}(\cdot|\theta_t)$. Further details on the network architecture are provided in Appendix \ref{app:exp_details}.  

\begin{figure*}[t!]
\centering
\begin{subfigure}{.325\textwidth}
  \centering
  \includegraphics[width=\linewidth]{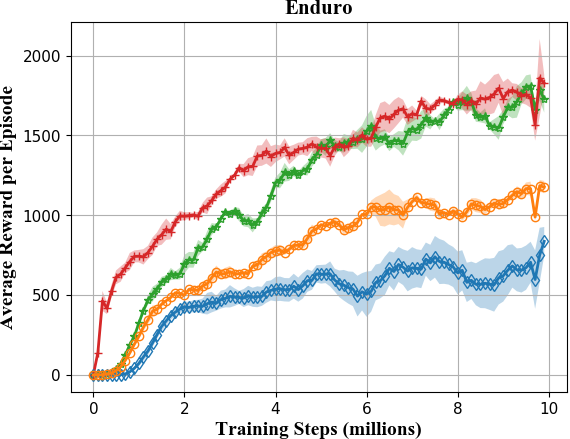}
\end{subfigure}%
\begin{subfigure}{.325\textwidth}
  \centering
  \includegraphics[width=\linewidth]{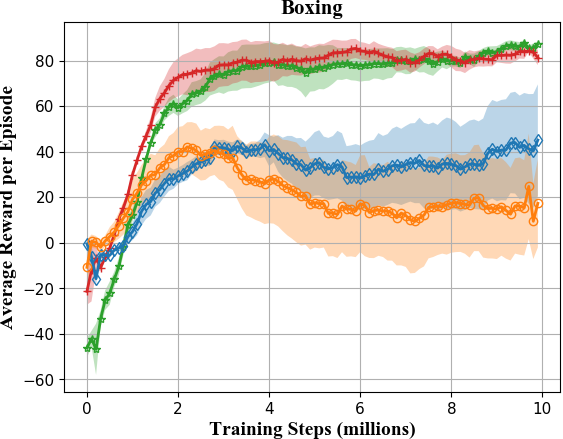}
\end{subfigure}
\begin{subfigure}{.325\textwidth}
  \centering
  \includegraphics[width=\linewidth]{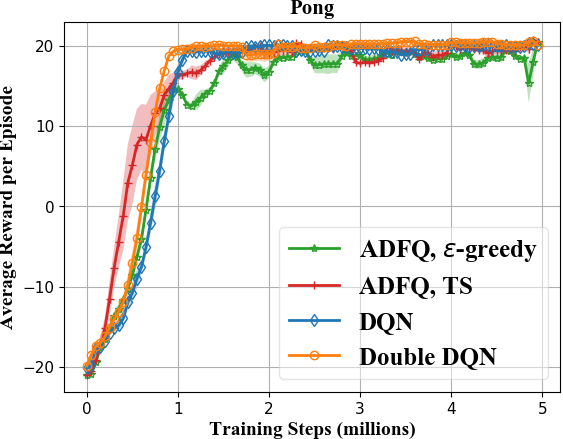}
\end{subfigure}\\
\begin{subfigure}{.325\textwidth}
  \centering
  \includegraphics[width=\linewidth]{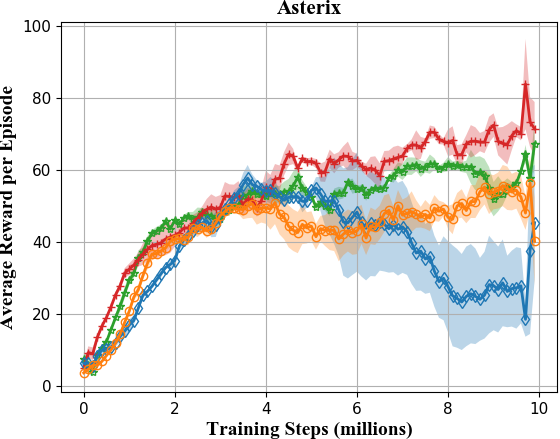}
\end{subfigure}
\begin{subfigure}{.325\textwidth}
  \centering
  \includegraphics[width=\linewidth]{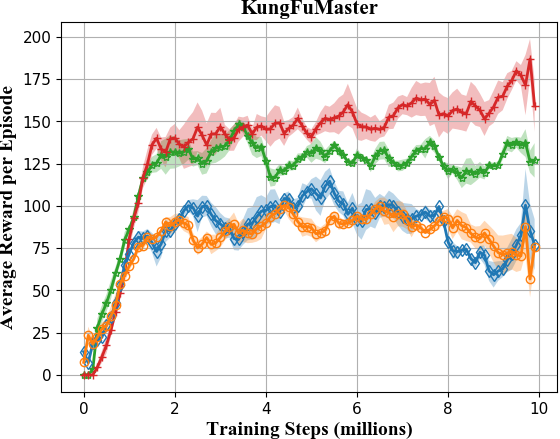}
\end{subfigure}
\begin{subfigure}{.325\textwidth}
  \centering
  \includegraphics[width=\linewidth]{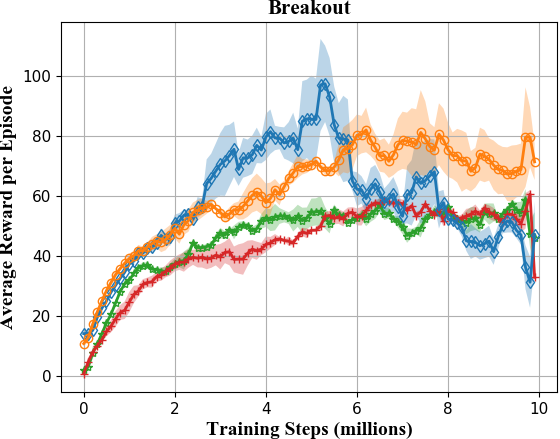}
\end{subfigure}
\caption{\footnotesize Performance of ADFQ, DQN, and Double DQN during learning smoothed by a moving average with window 6.}
\label{fig:deep_test}
\end{figure*}

The algorithms were evaluated for $T_H=10M$ training steps (5M for Pong). Each learning was greedily evaluated at every epoch ($=T_H/100$) for 3 times, and their averaged results are presented in Fig.\ref{fig:deep_test}. The entire experiment was repeated for 3 random seeds. Rewards were normalized to $\{-1,0,1\}$ and different from raw scores of the games. Both ADFQ with TS and with $\epsilon$-greedy notably surpassed DQN and Double DQN in Enduro, Boxing, Asterix, and Kung-Fu Master and showed similar results in Pong. The performance of ADFQ in Breakout is explained as Breakout is the only tested domain where there is no dynamic object interrupting the learning agent. As the demonstration in Sec.\ref{sec:demo} and the additional experiments in the appendix show, improvements of ADFQ from Q-learning is more significant when an experimental domain has high stochasticity and its action space is large due to the non-greedy update with uncertainty measures. Additionally, ADFQ showed more stable performance in all tested domains overcoming DQN's instability. ADFQ with TS achieved slightly higher performance than the $\epsilon$-greedy method utilizing the uncertainty in exploration. 

\section{Discussion} \label{sec:Discussion}
We proposed an approach to Bayesian off-policy TD method called ADFQ. ADFQ demonstrated that it could improve some of the issues from the greedy update of Q-learning by showing the quicker convergence to $Q^*$ than Q-learning and surpassing DQN and Double DQN in various Atari games. The presented ADFQ algorithm demonstrates several intriguing results. 

\textbf{Non-greedy update regularized by uncertainty measures.} Unlike the conventional Q-learning algorithm, ADFQ incorporates the information of all available actions for the subsequent state in the Q-value update. Each subsequent state-action pair contributes to the update based on its TD target mean and variance as well as its TD error. Particularly, we make use of our uncertainty measures not only in exploration but also in the value update as natural regularization. The advantages of the non-greedy update are noticeable in highly stochastic domains or domains with a large action space in the experiment.

\textbf{Convergence to Q-learning.} We prove that ADFQ converges to Q-learning as the variances decrease and can be seen as a more general form of Q-learning. 

\textbf{Computational Complexity and Scalability.} One of the major drawbacks of Baysian RL approaches is their high computational complexity \citep{Ghavamzadeh2015}. ADFQ is computationally efficient and is extended to Deep ADFQ with a neural network. 

We would like to highlight the fact that ADFQ is a Bayesian counterpart of Q-learning and is orthogonal to most other advancements made in Deep RL. Deep ADFQ merely changes the loss function and we compare with basic architectures here to provide insight as to how it may improve the performance. ADFQ can be used in conjunction with other extensions and techniques applied to Q-learning and DQN.

\bibliographystyle{abbrvnat}
\bibliography{main}
\newpage
\onecolumn
\begin{appendix}
\end{appendix}
\centerline{\LARGE \textbf{Appendix}}
\vspace{16pt}
\section{Maximum of Gaussian Random Variables}
\label{app:max_gaussian}
Let $X_i$ follows a Gaussian distribution, $\mathcal{N}(\mu_i, \sigma^2_i)$, and $\mu_i\neq \mu_j, \sigma_i \neq \sigma_j$ for any $i\neq j$. The distribution of the maximum of independent Gaussian random variables is derived as follows:
\begin{equation*}
    Pr\left(\max_{1\leq i \leq N} X_i \leq x \right) = \prod^N_{i=1}Pr(X_i \leq x) = \prod^N_{i=1} \Phi \left( \frac{x-\mu_i}{\sigma_i} \right)
\end{equation*}
\begin{align}
    p \left(\max_{1\leq i \leq N} X_i = x \right) =&\; \frac{d}{dx}\left(Pr\left(\max_{1\leq i \leq N} X_i \leq x \right) \right) \nonumber \\
    =&\; \sum^N_{i=1} \frac{1}{\sigma_i}\phi \left( \frac{x-\mu_i}{\sigma_i} \right) \prod^N_{i\neq j} \Phi \left( \frac{x-\mu_j}{\sigma_j} \right) \neq \text{\normalsize Gaussian} \label{eq:A_max_pdf}
\end{align}
where $\phi(\cdot)$ is the standard Gaussian probability density function (PDF) and $\Phi(\cdot)$ is the standard Gaussian cumulative distribution function (CDF).

\section{Derivation of the Posterior Distribution of $Q$} \label{app:posterior_derivation}
In the section \ref{sec:Qbelief} of the main paper, we have shown that 
\begin{equation*}
  \hat{p}_{Q_{s,a}}(q|\mathbf{\theta},r,s') = \frac{1}{Z} p_{ V_{s'}}\left( \left. \frac{q-r}{\gamma} \right| q,s',\mathbf{\theta}\right) p_{Q_{s,a}}(q|\mathbf{\theta})
\end{equation*}
where $Z$ is a normalization constant. Applying the distributions over $V_{s'}$ and $Q_{s,a}$, the posterior is derived as:
\begin{align*}
  \hat{p}_{Q_{s,a}}(q) =&\; \frac{1}{Z} \sum_{b\in \mathcal{A}} \frac{1}{ \sigma_{s',b}} \phi \left( \frac{q - (r + \gamma \mu_{s',b})}{\gamma\sigma_{s',b}} \right) \prod_{b'\neq b, b'\in \mathcal{A}} \Phi \left(\frac{q - (r + \gamma \mu_{s',b'})}{\gamma\sigma_{s',b'}}\right) \frac{1}{\sigma_{s,a}} \phi \left( \frac{q - \mu_{s,a}}{\sigma_{s,a}} \right) \nonumber \\
  =&\; \frac{1}{Z\sqrt{2 \pi}\sigma_{s,a}} \sum_{b\in \mathcal{A}} \frac{1}{\sigma_{s',b}} \exp\left\{-\frac{1}{2} \frac{(\mu_{s,a} - (r+\gamma \mu_{s',b}))^2}{\sigma_{s,a}^2 + \gamma^2\sigma_{s',b}^2} \right\} \phi \left( \frac{q - \bar{\mu}_{\tau,b}}{\bar{\sigma}_{\tau,b}} \right) \nonumber \\
  &\; \hspace{200pt} \times \prod_{b'\neq b, b'\in \mathcal{A}} \Phi \left(\frac{q - (r + \gamma \mu_{s',b'})}{\gamma\sigma_{s',b'}}\right)\nonumber \\
  =&\; \frac{1}{Z} \sum_{b\in \mathcal{A}} \frac{c_{\tau,b}}{\bar{\sigma}_{\tau,b}} \phi \left( \frac{q - \bar{\mu}_{\tau,b}}{\bar{\sigma}_{\tau,b}} \right) \prod_{b'\neq b, b'\in \mathcal{A}} \Phi \left(\frac{q - (r + \gamma \mu_{s',b'})}{\gamma\sigma_{s',b'}}\right)
\end{align*}
where $Z$ is a normalization constant and 
\begin{equation*}
  c_{\mathbf{\tau},b} = \frac{1}{\sqrt{\sigma_{s,a}^2 + \gamma^2\sigma_{s',b}^2 }} \phi \left(\frac{(r+\gamma \mu_{s',b})- \mu_{s,a}}{\sqrt{\sigma_{s,a}^2 + \gamma^2\sigma_{s',b}^2}} \right) \label{eq:A_weight}
\end{equation*}
\begin{equation*}
  \bar{\mu}_{\mathbf{\tau},b} = \bar{\sigma}^2_{\mathbf{\tau},b} \Big(\frac{\mu_{s,a}}{\sigma_{s,a}^2} + \frac{r+ \gamma \mu_{s',b}}{\gamma^2 \sigma_{s',b}^2} \Big) \qquad \bar{\sigma}^2_{\mathbf{\tau},b} = \Big(\frac{1}{\sigma_{s,a}^2} + \frac{1}{\gamma^2\sigma_{s',b}^2} \Big)^{-1} \label{eq:A_var_bar}
\end{equation*}

\section{Mean and Variance of the Posterior Distribution of $Q$} \label{app:mgf}
\subsection{Moment Generating Function}
The mean and variance of the posterior distribution (Eq.\ref{eq:posterior}) can be analytically found when $|\mathcal{A}|=2$. Consider a random variable $X_M = \max_{1\leq k \leq N} X_k$ which density function (Eq.\ref{eq:A_max_pdf}) has a similar form to the posterior distribution. The moment generating function of $X_M$ is:
\begin{align*}
  M(t) =&\; \int^{\infty}_{-\infty} e^{tx} \sum_i \frac{1}{\sigma_i} \phi \Big(\frac{x - \mu_i}{\sigma_i} \Big) \prod_{i\neq j}\Phi\Big( \frac{x-\mu_j}{\sigma_j} \Big) dx\\
  =&\; \sum_i \eta_i(t) \int^{\infty}_{-\infty} \frac{1}{\sigma_i} \phi \bigg( \frac{x-\mu_i'}{\sigma_i} \bigg) \prod_{i\neq j}\Phi \bigg( \frac{x-\mu_j}{\sigma_j}\bigg) dx
\end{align*}
where 
\begin{equation*}
  \eta_i(t) = \exp\left\{\mu_i t+\frac{t^2 \sigma_i^2}{2}\right\} \qquad \text{and} \qquad \mu_i' = \mu_i + t\sigma_i^2
\end{equation*}
When $N=2$,
\begin{equation}
  M(t) = \int^{\infty}_{-\infty} e^{tx} \Bigg( \frac{1}{\sigma_1} \phi \Big(\frac{x - \mu_1}{\sigma_1} \Big) \Phi\Big( \frac{x-\mu_2}{\sigma_2} \Big) + \frac{1}{\sigma_2} \phi \Big(\frac{x - \mu_2}{\sigma_2} \Big) \Phi\Big( \frac{x-\mu_1}{\sigma_1} \Big) \Bigg) dx
\end{equation}
Since the two terms are symmetric, let $M(t) = M_1(t) + M_2(t)$ and differentiate each term with respect to $\mu_2$ and $\mu_1$, respectively. For the first term,
\begin{align}
  \frac{ \partial M_1(t)}{\partial \mu_2} =&\; - \frac{\eta_1(t)}{\sigma_1\sigma_2} \int^{\infty}_{-\infty} \phi\Big(\frac{x-\mu_1'}{\sigma_1} \Big)\phi\Big(\frac{x-\mu_2}{\sigma_2} \Big) dx \nonumber \\
  =&\; -\frac{\eta_1(t)\sigma_{12}}{\sqrt{2\pi} \sigma_1 \sigma_2} \exp\left\{ -\frac{1}{2}\frac{(\mu_1'-\mu_2)^2}{\sigma_1^2 + \sigma_2^2} \right\} \int^{\infty}_{-\infty} \frac{1}{\sigma_{12}}\phi \left(\frac{x-\mu_{12}}{\sigma_{12}} \right) dx \nonumber\\
  =&\; -\frac{\eta_1(t)\sigma_{12}}{\sqrt{2\pi} \sigma_1 \sigma_2} \exp\left\{ -\frac{1}{2}\frac{(\mu_1'-\mu_2)^2}{\sigma_1^2 + \sigma_2^2} \right\} \label{eq:A_m1_derivative}
\end{align}
where
\[
	\mu_{12} = \sigma_{12}^2\left(\frac{\mu_1'}{\sigma_1^2} + \frac{\mu_2}{\sigma_2^2} \right) \qquad \qquad \frac{1}{\sigma_{12}^2} = \frac{1}{\sigma_1^2} + \frac{1}{\sigma_2^2}
\]
Then, we integrate Eq.\ref{eq:A_m1_derivative} with respect to $\mu_2$, 
\begin{align}
  M_1(t) =&\; \int \frac{\partial M_1(t)}{\partial \mu_2} d\mu_2 \nonumber\\
  =&\; - \frac{\eta_1(t)\sigma_{12}}{\sigma_1 \sigma_2} \sqrt{\sigma_1^2 +\sigma_2^2} \int \frac{1}{ \sqrt{ 2 \pi (\sigma_1^2 + \sigma_2^2)}} \exp \left\{ -\frac{(\mu_1' - \mu_2)^2}{2(\sigma_1^2 + \sigma_2^2)}  d \mu_2\right\}\nonumber \\
  =&\; \eta_1(t) \Phi \Bigg( \frac{\mu_1' - \mu_2}{\sqrt{\sigma_1^2 + \sigma_2^2}} \Bigg) \label{eq:moment_1}
\end{align}
\subsection{Moments of the Posterior Distribution}
We apply the result in Eq.\ref{eq:moment_1} to the posterior distribution by replacing the variables in $M_1(t)$ as:
\[
	\mu_1 \rightarrow \bar{\mu}_{\tau,1} \quad \sigma_1 \rightarrow \bar{\sigma}_{\tau,1} \quad \mu_2 \rightarrow r+\gamma \mu_2\quad \sigma_2 \rightarrow \gamma \sigma_2
\]
and replacing the variables in $M_2(t)$ similarly. Then, we obtain the normalizing factor:
\begin{equation}
Z = c_{\tau,1}\Phi_{\tau,1} + c_{\tau,2}\Phi_{\tau,2} \label{eq:A_normalization}
\end{equation}
where we define the following notations for simplicity:
\[
\Phi_{\tau,1} \equiv  \Phi\left( \frac{\bar{\mu}_{\tau,1}-(r+\gamma \mu_{s',2})}{\sqrt{\bar{\sigma}_{\tau,1}^2 + \gamma^2 \sigma_{s',2}^2}} \right),\qquad 
\phi_{\tau,1} \equiv \frac{1}{\sqrt{\bar{\sigma}_{\tau,1}^2 + \gamma^2 \sigma_{s',2}^2}} \phi\left( \frac{\bar{\mu}_{\tau,1}-(r+\gamma \mu_{s',2})}{\sqrt{\bar{\sigma}_{\tau,1}^2 + \gamma^2 \sigma_{s',2}^2}}  \right)
\] 
and $\Phi_{\tau,2}$ and $\phi_{\tau,2}$ are also similarly defined. The exact mean of the posterior distribution is derived by solving the first derivative of the moment generating function with respect to $t$ at $t=0$:
\begin{equation*}
M_1'(t) = c_{\tau,1}\eta_1'(t)  \Phi\left( \frac{\bar{\mu}_{\tau,1}'-(r+\gamma \mu_{s',2})}{\sqrt{\bar{\sigma}_{\tau,1}^2 + \gamma^2 \sigma_{s',2}^2}} \right) + c_{\tau,1}\eta_1(t) \frac{\bar{\sigma}_{\tau,1}^2}{\sqrt{\bar{\sigma}_{\tau,1}^2 + \gamma^2 \sigma_{s',2}^2}} \phi\left( \frac{\bar{\mu}_{\tau,1}' -(r+\gamma \mu_{s',2})}{\sqrt{\bar{\sigma}_{\tau,1}^2 + \gamma^2 \sigma_{s',2}^2}}\right)
\end{equation*}
\begin{align}
 \E_{q \sim \hat{p}_{Q_{s,a}}(\cdot)}[q] =&\; \frac{1}{Z}\left(M'_1(t=0) + M'_2(t=0) \right) \nonumber \\
 =&\; \frac{c_{\tau,1}}{Z}\left( \bar{\mu}_{\tau,1} \Phi_{\tau,1} + \bar{\sigma}_{\tau,1}^2 \phi_{\tau,1} \right)  +  \frac{c_{\tau,2}}{Z} \left( \bar{\mu}_{\tau,2} \Phi_{\tau,2} + \bar{\sigma}_{\tau,2}^2 \phi_{\tau,2} \right)  \label{eq:A_exact_1st_moment}
 \end{align}
The variance of the posterior is also derived by solving the second derivative of the moment generating function:
\begin{align*}
M_1''(t) =&\; c_{\tau,1}\eta_1''(t)  \Phi\left( \frac{\bar{\mu}_{\tau,1}'-(r+\gamma \mu_{s',2})}{\sqrt{\bar{\sigma}_{\tau,1}^2 + \gamma^2 \sigma_{s',2}^2}} \right) + 2c_{\tau,1}\eta_1'(t) \frac{\bar{\sigma}_{\tau,1}^2}{\sqrt{\bar{\sigma}_{\tau,1}^2 + \gamma^2 \sigma_{s',2}^2}} \phi\left( \frac{\bar{\mu}_{\tau,1}' -(r+\gamma \mu_{s',2})}{\sqrt{\bar{\sigma}_{\tau,1}^2 + \gamma^2 \sigma_{s',2}^2}}\right)\\
&\; \qquad + c_{\tau,1} \eta_1(t) \left(-\frac{\bar{\mu}_{\tau,1}'-(r+\gamma \mu_{s',2})}{(\bar{\sigma}_{\tau,1}^2 + \gamma^2 \sigma_{s',2}^2)\bar{\sigma}_{\tau,1}^{-2}} \right)  \frac{\bar{\sigma}_{\tau,1}^2}{\sqrt{\bar{\sigma}_{\tau,1}^2 + \gamma^2 \sigma_{s',2}^2}} \phi\left( \frac{\bar{\mu}_{\tau,1}' -(r+\gamma \mu_{s',2})}{\sqrt{\bar{\sigma}_{\tau,1}^2 + \gamma^2 \sigma_{s',2}^2}}\right)
\end{align*}
Thus, the second moment is:
{\small
\begin{align}
	&\;\E_{q \sim \hat{p}_{Q_{s,a}}(\cdot)}[q^2] = \frac{c_{\tau,1}}{Z}\left( (\bar{\mu}_{\tau,1}^2 + \bar{\sigma}_{\tau,1}^2)\Phi_{\tau,1} + 2\bar{\mu}_{\tau,1}\bar{\sigma}_{\tau,1}^2 \phi_{\tau,1} - \frac{\bar{\sigma}_{\tau,1}^4}{\bar{\sigma}_{\tau,1}^2 + \gamma^2 \sigma_{s',2}^2}(\bar{\mu}_{\tau,1}-(r+\gamma \mu_{s',2}))\phi_{\tau,1} \right) \nonumber\\
    &\; \quad \frac{c_{\tau,2}}{Z}\left( (\bar{\mu}_{\tau,2}^2 + \bar{\sigma}_{\tau,2}^2)\Phi_{\tau,2} + 2\bar{\mu}_{\tau,2}\bar{\sigma}_{\tau,2}^2 \phi_{\tau,2} - \frac{\bar{\sigma}_{\tau,2}^4}{\bar{\sigma}_{\tau,2}^2 + \gamma^2 \sigma_{s',1}^2}(\bar{\mu}_{\tau,2}-(r+\gamma \mu_{s',1}))\phi_{\tau,2} \right) \label{eq:A_exact_2nd_moment}
\end{align}}
and the variance is $\E_{q \sim \hat{p}_{Q_{s,a}}(\cdot)}[q^2] -( \E_{q \sim \hat{p}_{Q_{s,a}}(\cdot)}[q])^2$.

\begin{figure}[tb!]
\centering
        \includegraphics[width =0.45\textwidth]{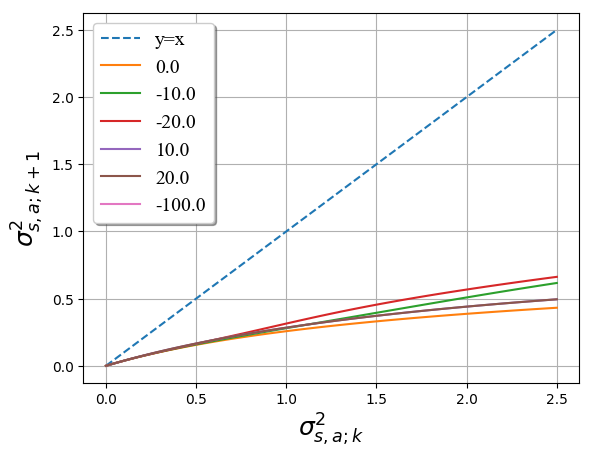}
  \qquad 
        \includegraphics[width =0.45\textwidth]{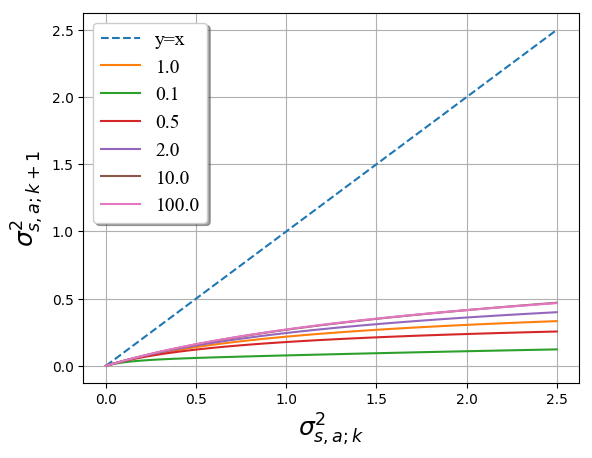}
  \caption{\footnotesize{Relationship between $\sigma_{s, a;k}^2$ and its updated value, $\sigma_{s,a;k+1}^2$ for $|\mathcal{A}|=2$. Each solid curve represents a different set of parameters. Left: Differing values of $\mu_{s^\prime, 2} - \mu_{s^\prime, 1}$. Right: Differing values of $\sigma_{s^\prime, 1}^2 / \sigma_{s^\prime, 2}^2$    }}
  \label{fig:var_decrease}
\end{figure}

Fig.\ref{fig:var_decrease} shows empirical evidence of the contraction behavior of the variance update, $\Var_{q \sim \hat{p}_{Q_{s,a}}(\cdot)}[q] < \sigma_{s,a}^2$ which is one of the conditions for Theorem \ref{theorem:convergence_numeric}. The updated variance is less than the current variance for a large range of different values for the related parameters. In addition, it is easily shown that 0 is the fixed point of the variance from Eq.\ref{eq:A_exact_1st_moment}-\ref{eq:A_exact_2nd_moment}.

\section{Q-beliefs with Gaussian white noise} \label{app:stochastic_mdp}
In order to incorporate stochasticity of an MDP, we add small Gaussian white noise to the likelihood, $r+\gamma V_{s'} + W$ where $W \sim \mathcal{N}(0,\sigma^2_w)$, and the likelihood distribution is obtained by solving the following integral:
\begin{align}
p(r+\gamma V_{s'}|q,\theta) = &\; \int_{-\infty}^{\infty} \sum_{b\in \mathcal{A}} \frac{1}{\sigma_{s',b}} \phi \left( \frac{w-(q-(r+\gamma\mu_{s',b}))}{\gamma \sigma_{s',b}} \right) \nonumber \\
&\; \qquad \qquad \times \prod_{b' \neq b} \Phi \left( - \frac{w-(q-(r+\gamma \mu_{s',b'}))}{\gamma \sigma_{s',b'}} \right) \frac{1}{\sigma_{w}} \phi\left( \frac{w}{\sigma_w} \right) dw \nonumber\\
=&\;  \int_{-\infty}^{\infty}  \sum_{b\in\mathcal{A}} \frac{l_b}{\bar{v}_b} \phi \left( \frac{w-\bar{w}_b}{\bar{v}_b} \right) \prod_{b'\neq b}  \left( 1 - \Phi\left(\frac{w-(q-r-\gamma \mu_{s',b'})}{\gamma \sigma_{s',b'}} \right) \right) dw \label{eq:expected_likelihood}
\end{align}
where 
\begin{equation}
l_b = \frac{1}{\sqrt{\sigma_w^2 + \gamma^2 \sigma^2_{s',b}}} \phi \left( \frac{q-(r+\gamma \mu_{s',b})}{\sqrt{\sigma_w^2 + \gamma^2 \sigma^2_{s',b}}} \right)
\end{equation}
\begin{equation}
\bar{w}_b = \bar{v}_b^2 \left( \frac{q-(r+\gamma \mu_{s',b})}{\gamma^2 \sigma^2_{s',b}} \right) \qquad \qquad \frac{1}{\bar{v}^2_b} = \frac{1}{\gamma^2 \sigma^2_{s',b}} + \frac{1}{\sigma_w^2} \label{eq:sto_bars}
\end{equation}

\subsection{Expected Likelihood for $|\mathcal{A}|=2$}
The distribution inside the integral in Eq.\ref{eq:expected_likelihood} has a similar form with the posterior distribution Eq.\ref{eq:posterior}. As mentioned above, a closed form solution for its integral is not available when $|\mathcal{A}|>2$. Therefore, we derive an analytic solution of the expected likelihood when $|\mathcal{A}|=2$ and approximate to a simpler form so that it can be generalized to an arbitrary number of actions.

Using Eq.\ref{eq:moment_1} for finding the zeroth moment, we obtain:
\begin{equation}
p(r+\gamma V_{s'}|q,\theta) = l_1 \Phi \left( - \frac{\bar{w}_1-(q-(r+\gamma \mu_{s',2}))}{\sqrt{\bar{v}_1^2 + \gamma^2 \sigma_{s',2}^2}} \right) + l_2 \Phi \left( - \frac{\bar{w}_2-(q-(r+\gamma \mu_{s',1}))}{\sqrt{\bar{v}_2^2 + \gamma^2 \sigma_{s',1}^2}} \right) \label{eq:sol_ll_2act}
\end{equation}
Inside the CDF term is a function of $q$:
{\small
\begin{equation*}
 - \frac{\bar{w}_1-(q-(r+\gamma \mu_{s',2}))}{\sqrt{\bar{v}_1^2 + \gamma^2 \sigma_{s',2}^2}} = \frac{1}{\sqrt{\bar{v}_1^2 + \gamma^2 \sigma_{s',2}^2}}\left( \left(1 - \frac{\bar{v}_1^2}{\gamma^2\sigma_{s',1}^2}\right) q - \left( r+\gamma \mu_{s',2} - \frac{\bar{v}_1^2}{\gamma^2\sigma_{s',1}^2} (r+\gamma\mu_{s',1}) \right)\right)
\end{equation*}}
We define 
\[
	 \mu^w_2 \equiv  \left(1 - \frac{\bar{v}_1^2}{\gamma^2\sigma_{s',1}^2}\right)^{-1}\left( r+\gamma \mu_{s',2} - \frac{\bar{v}_1^2}{\gamma^2\sigma_{s',1}^2} (r+\gamma\mu_{s',1}) \right)
\]
\[
	\sigma^w_2 \equiv  \left(1 - \frac{\bar{v}_1^2}{\gamma^2\sigma_{s',1}^2}\right)^{-1} \sqrt{\bar{v}_1^2 + \gamma^2 \sigma_{s',2}^2}
\]
and express the likelihood distribution Eq.\ref{eq:sol_ll_2act} as: 
\begin{equation}
p(r+\gamma V_{s'}|q,\theta) = l_1  \Phi \left(\frac{q-\mu^w_2}{\sigma^w_2} \right)+ l_2 \Phi \left(\frac{q-\mu^w_1}{\sigma^w_1}  \right) \label{eq:likelihood_sto}
\end{equation}
Then, we can find the solutions of the posterior mean and variance for $\sigma_w>0$ when $|\mathcal{A}|=2$ by replacing $r+\gamma\mu_{s',2}$ and $\gamma \sigma_{s',2}$ with $\mu_2^w$ and $\sigma_2^w$, respectively in Eq.\ref{eq:A_exact_1st_moment} and Eq.\ref{eq:A_exact_2nd_moment}.

\subsection{Asymptotic Limits} \label{app:sto_asymp}
In one asymptotic limit of $\sigma_w/\sigma_{s',b} \rightarrow 0$, 
\[
	\lim_{\sigma_w/\sigma_{s',b} \rightarrow 0} \bar{v}_b^2 + \gamma^2 \sigma^2_{s',b'} = \lim_{\sigma_w/\sigma_{s',b} \rightarrow 0} \frac{\gamma^2 \sigma^2_{s',b} \sigma_w^2}{\gamma^2 \sigma^2_{s',b} + \sigma^2_w}+ \gamma^2 \sigma^2_{s',b'} =  \gamma^2 \sigma^2_{s',b'}
\]
\[
 \lim_{\sigma_w/\sigma_{s',b} \rightarrow 0} \frac{\bar{v}_b^2}{\gamma^2 \sigma_{s',b}^2} = \lim_{\sigma_w/\sigma_{s',b} \rightarrow 0} \frac{ \sigma_w^2}{\gamma^2 \sigma^2_{s',b} + \sigma^2_w} = 0
\]
and therefore,
\begin{equation}
	\lim_{\sigma_w/\sigma_{s',b} \rightarrow 0} \Phi \left( - \frac{\bar{w}_b-(q-(r+\gamma \mu_{s',b'}))}{\sqrt{\bar{v}_b^2 + \gamma^2 \sigma_{s',b'}^2}} \right) = \Phi \left(\frac{q-(r+\gamma\mu_{s',b'})}{\gamma\sigma_{s',b'}} \right) 
\end{equation}
and the likelihood distribution becomes 
\begin{equation}
 \sum_{b\in \{1,2\}} \frac{1}{\sqrt{\sigma_w^2 + \gamma^2 \sigma^2_{s',b}}} \phi \left( \frac{q-(r+\gamma \mu_{s',b})}{\sqrt{\sigma_w^2 + \gamma^2 \sigma^2_{s',b}}} \right)\Phi \left(\frac{q-(r+\gamma\mu_{s',b'})}{\gamma\sigma_{s',b'}} \right) \label{eq:likelihood_sto_limit0}
\end{equation}
Therefore, the posterior distribution derived from this likelihood has the same form with Eq.\ref{eq:posterior} but it uses $\gamma^2 \sigma^2_{s',b}+\sigma_w^2$ instead of $\gamma^2 \sigma^2_{s',b}$ in $c_{\tau,b}$, $\bar{\mu}_{\tau,b}$, and $\bar{\sigma}_{\tau,b}$:
\begin{equation}
  c_{\mathbf{\tau},b} = \frac{1}{\sqrt{\sigma_{s,a}^2 + \gamma^2\sigma_{s',b}^2+ \sigma_w^2}} \phi \left(\frac{(r+\gamma \mu_{s',b})- \mu_{s,a}}{\sqrt{\sigma_{s,a}^2 + \gamma^2\sigma_{s',b}^2+ \sigma_w^2}} \right) \label{eq:A_weight_sto}
\end{equation}
\begin{equation}
  \bar{\mu}_{\mathbf{\tau},b} = \bar{\sigma}^2_{\mathbf{\tau},b} \Big(\frac{\mu_{s,a}}{\sigma_{s,a}^2} + \frac{r+ \gamma \mu_{s',b}}{\gamma^2 \sigma_{s',b}^2+ \sigma_w^2} \Big) \qquad \bar{\sigma}^2_{\mathbf{\tau},b} = \Big(\frac{1}{\sigma_{s,a}^2} + \frac{1}{\gamma^2\sigma_{s',b}^2+ \sigma_w^2} \Big)^{-1}\label{eq:A_bar_sto}
\end{equation}
It is identical to the posterior of the case when $\sigma_w = 0$. 

In the other asymptotic limit,
\[
	\lim_{\sigma_{s',b}/\sigma_w \rightarrow 0} \bar{v}_b^2 + \gamma^2 \sigma^2_{s',b'}  =  \gamma^2 \sigma^2_{s',b} + \gamma^2 \sigma^2_{s',b'}
\quad \text{  and  } \quad
 \lim_{\sigma_{s',b}/\sigma_w \rightarrow 0} \frac{\bar{v}_b^2}{\gamma^2 \sigma_{s',b}^2}  = 1
\]
Since we set $\sigma_w$ as a small number, $\sigma_{s',b}/\sigma_w \rightarrow 0$ infers $\sigma_{s',b} \rightarrow 0$, and therefore, the likelihood distribution becomes Gaussian :
\begin{align}
\lim_{\sigma_{s',b}/\sigma_w \rightarrow 0} p(r+\gamma V_{s'}|q,\theta) =&\; \lim_{\sigma_{s',b}/\sigma_w \rightarrow 0} l_1  \Phi \left(\frac{\mu_{s',1}-\mu_{s',2}}{\sqrt{\gamma^2 \sigma^2_{s',1} + \gamma^2 \sigma^2_{s',2}}} \right)+ l_2 \Phi \left(\frac{\mu_{s',2}-\mu_{s',1}}{\sqrt{\gamma^2 \sigma^2_{s',1} + \gamma^2 \sigma^2_{s',2}}}  \right) \nonumber \\
=&\; \frac{1}{\sqrt{\sigma_w^2 + \gamma^2 \sigma^2_{s',b^+}}} \phi \left( \frac{q-(r+\gamma \mu_{s',b^+})}{\sqrt{\sigma_w^2 + \gamma^2 \sigma^2_{s',b^+}}} \right) \label{eq:likelihood_sto_limit1}
\end{align}
where $b^+ = \argmax_{b\in \{1,2 \}} \mu_{s',b}$. Therefore, the posterior distribution becomes Gaussian with mean at $\bar{\mu}_{\tau,b^+}$ and variance at $\bar{\sigma}^2_{\tau,b^+}$ in Eq.\ref{eq:A_bar_sto}. 

\subsection{Approximate Likelihood}
In order to have closed-form expressions for the ADFQ update, we extend the asymptotic result for $|\mathcal{A}|=2$ presented in the previous section to the general case ($|\mathcal{A}|=n$ for $n \in \mathbb{N}$) with an assumption of $\sigma_w \ll \sigma_{s',b}$ $\forall b\in \mathcal{A}$. Therefore, the approximate likelihood is:
\begin{equation*}
 p(r+\gamma V_{s'}|q, \theta) = \sum_{b\in \mathcal{A}} \frac{\gamma}{ \sqrt{\gamma^2\sigma_{s',b}^2+\sigma^2_w}} \phi \left( \frac{q - (r + \gamma \mu_{s',b})}{\sqrt{\gamma^2\sigma_{s',b}^2+\sigma^2_w}} \right) \prod_{b'\neq b, b'\in \mathcal{A}} \Phi \left(\frac{q - (r + \gamma \mu_{s',b'})}{\gamma\sigma_{s',b'}}\right)
\end{equation*} 
Then, the posterior distribution is derived as:
\begin{align*}
  \hat{p}_{Q_{s,a}}(q) =&\; \frac{1}{Z} \sum_{b\in \mathcal{A}} \frac{\gamma}{ \sqrt{\gamma^2\sigma_{s',b}^2+\sigma^2_w}} \phi \left( \frac{q - (r + \gamma \mu_{s',b})}{\sqrt{\gamma^2\sigma_{s',b}^2+\sigma^2_w}} \right)\\
  &\; \hspace{120pt} \times \prod_{b'\neq b, b'\in \mathcal{A}} \Phi \left(\frac{q - (r + \gamma \mu_{s',b'})}{\gamma\sigma_{s',b'}}\right) \frac{1}{\sigma_{s,a}} \phi \left( \frac{q - \mu_{s,a}}{\sigma_{s,a}} \right)  \\
  =&\; \frac{1}{Z\sqrt{2 \pi}\sigma_{s,a}} \sum_{b\in \mathcal{A}} \frac{\gamma}{ \sqrt{\gamma^2\sigma_{s',b}^2+\sigma^2_{\epsilon}}} \exp\left\{-\frac{1}{2} \frac{(\mu_{s,a} - (r+\gamma \mu_{s',b}))^2}{\sigma_{s,a}^2 + \gamma^2\sigma_{s',b}^2 + \sigma_w^2} \right\}  \\
  &\; \hspace{120pt} \times \phi \left( \frac{q - \bar{\mu}_{\tau,b}}{\bar{\sigma}_{\tau,b}} \right) \prod_{b'\neq b, b'\in \mathcal{A}} \Phi \left(\frac{q - (r + \gamma \mu_{s',b'})}{\gamma\sigma_{s',b'}}\right)\nonumber \\
  =&\; \frac{1}{Z} \sum_{b\in \mathcal{A}} \frac{c_{\tau,b}}{\bar{\sigma}_{\tau,b}} \phi \left( \frac{q - \bar{\mu}_{\tau,b}}{\bar{\sigma}_{\tau,b}} \right) \prod_{b'\neq b, b'\in \mathcal{A}} \Phi \left(\frac{q - (r + \gamma \mu_{s',b'})}{\gamma\sigma_{s',b'}}\right)
\end{align*}
where $Z$ is a normalization constant and 
\begin{equation*}
  c_{\mathbf{\tau},b} = \frac{1}{\sqrt{\sigma_{s,a}^2 + \gamma^2\sigma_{s',b}^2+ \sigma_w^2}} \phi \left(\frac{(r+\gamma \mu_{s',b})- \mu_{s,a}}{\sqrt{\sigma_{s,a}^2 + \gamma^2\sigma_{s',b}^2+ \sigma_w^2}} \right) 
\end{equation*}
\begin{equation*}
  \bar{\mu}_{\mathbf{\tau},b} = \bar{\sigma}^2_{\mathbf{\tau},b} \Big(\frac{\mu_{s,a}}{\sigma_{s,a}^2} + \frac{r+ \gamma \mu_{s',b}}{\gamma^2 \sigma_{s',b}^2+ \sigma_w^2} \Big) \qquad \bar{\sigma}^2_{\mathbf{\tau},b} = \Big(\frac{1}{\sigma_{s,a}^2} + \frac{1}{\gamma^2\sigma_{s',b}^2+ \sigma_w^2} \Big)^{-1}
\end{equation*}

\section{Proofs} \label{app:proofs}
\subsection{Lemma 1}
\textbf{Lemma 1.}\textit{
Let $X$ be a random variable following a normal distribution, $\mathcal{N}(\mu,\sigma^2)$. Then we have:}
\begin{equation}
	\lim_{\sigma \rightarrow 0 } \left[\Phi \left(\frac{x-\mu}{\sigma}\right) - \exp\left\{ -\frac{1}{2}\left[ -\frac{x-\mu}{\sigma} \right]_+^2 \right\} \right] = 0
\end{equation}
\textit{where $[x]_+=\max(0,x)$ is the {\em ReLU} nonlinearity.}
\begin{proof}
\[
	\lim_{\sigma \rightarrow 0} \frac{x-\mu}{\sigma} = -\infty
\]
Let's define $y \equiv (x-\mu)/\sigma$,
\begin{align*}
\Phi(y< 0) =&\; \int^y_{-\infty} \frac{1}{\sqrt{2\pi}}e^{-\frac{1}{2}t^2} dt \\
=&\; \int^0_{-\infty} \frac{1}{\sqrt{2\pi}}\exp\left\{-\frac{1}{2}(y+t')^2 \right\} dt'\\
=&\; \frac{1}{\sqrt{2\pi}} \exp\left\{-\frac{1}{2}y^2\right\} \int^0_{-\infty} \exp\left\{-\left(y+\frac{t'}{2}\right)t' \right\} dt'\\
= &\; \frac{1}{2} \exp\left\{-\frac{1}{2}y^2\right\} \int^0_{-\infty} \exp\left\{-yt' \right\} dt'\\
= &\; -\frac{1}{2y}\exp\left\{-\frac{1}{2}y^2 \right\} \\
= &\; \frac{1}{2|y|}\exp\left\{-\frac{1}{2}y^2 \right\} \\
\end{align*}
\begin{align*}
	\lim_{y<0, y\rightarrow -\infty} \left[ \Phi(y) - \exp\left\{-\frac{1}{2}y^2 \right\} \right] = \lim_{y<0, y\rightarrow -\infty} \left( 1-\frac{1}{2|y|}\right) \exp\left\{-\frac{1}{2}y^2 \right\} = 0
\end{align*}
For $x \geq \mu$ and $y \geq 0$ and 
\[
	\lim_{\sigma \rightarrow 0} \frac{x-\mu}{\sigma} = \infty
\]
\[
	\lim_{y\rightarrow \infty} \Phi(y) =\lim_{y\rightarrow \infty}  \int^y_{-\infty} \frac{1}{\sqrt{2\pi}}e^{-\frac{1}{2}t^2} dt = 1 = e^0
\]
Therefore, 
\begin{align*}
	\lim_{\sigma \rightarrow 0 } \left[\Phi \left(\frac{x-\mu}{\sigma}\right) - \exp\left\{ -\frac{1}{2}\left[ -\frac{x-\mu}{\sigma} \right]_+^2 \right\} \right] = 0
\end{align*}
\end{proof}
\subsection{Theorem 1}
\textbf{Theorem 1.}\textit{
Suppose that the mean and variance of $Q_{s,a}$ $\forall s \in \mathcal{S}, \forall a \in \mathcal{A}$ are iteratively updated by the mean and variance of $\hat{p}_{Q_{s,a}}$ after observing $r$ and $s'$ at every step. When $|\mathcal{A}|=2$, the update rule of the means is equivalent to the Q-learning update if all state-action pairs are visited infinitely often and the variances approach 0.
In other words, at the $k$th update on $\mu_{s,a}$:
\[
	\lim_{k\rightarrow \infty, \{\sigma\} \rightarrow 0}\mu_{s,a;k+1} = \left(1-\alpha_{\tau;k} \right) \mu_{s,a;k} + \alpha_{\tau;k} \big(r+\gamma \max_{b\in \mathcal{A}} \mu_{s',b;k}\big)
   \]
 where $\alpha_{\tau;k}=\sigma_{s,a;k}^2 / \left( \sigma_{s,a;k}^2 + \gamma^2 \sigma_{s',b^+;k}^2 + \sigma_w^2 \right)$ and $b^+ = \argmax_{b\in\mathcal{A}} \mu_{s',b}$.}
\begin{proof}
For simplicity, we first show the convergence of the algorithm for $\sigma_w=0$ and then extend the result to the general case.

For simplicity, we define new notations as: 
\[
   y_b \equiv r+\gamma \mu_{s',b}-\mu_{s,a}, \qquad 
   v_0 \equiv \sigma^2_{s,a}, \qquad 
   v_b \equiv \sigma_{s,a}^2 + \gamma^2 \sigma_{s',b}^2
\]
In the section \ref{app:mgf}, we obtained the exact solutions for the posterior mean and variance when $|\mathcal{A}|=2$ (Eq.\ref{eq:A_exact_1st_moment} and Eq.\ref{eq:A_exact_2nd_moment}).
When $\sigma_{s,a}, \sigma_{s',a_1}, \sigma_{s',a_2} \rightarrow 0$, the posterior mean is approximated as: 
\begin{equation}
 \frac{\bar{\mu}_{\tau,1}c_{\tau,1}\Phi_{\tau,1} + \bar{\mu}_{\tau,2}c_{\tau,2}\Phi_{\tau,2}}{c_{\tau,1}\Phi_{\tau,1} + c_{\tau,2}\Phi_{\tau,2}} \label{eq:mean_approx}
 \end{equation}
 Then, using the Lemma.\ref{lemma:normal_approx}, $c_{\tau,1}\Phi_{\tau,1}$ is approximated as: 
 \begin{align}
 &\;  \frac{1}{\sqrt{2\pi (\sigma_{s,a}^2 + \gamma^2 \sigma_{s',1}^2)}}\exp\left\{ -\frac{(r+\gamma \mu_{s',1}-\mu_{s,a})^2}{2(\sigma_{s,a}^2 + \gamma^2 \sigma_{s',1}^2)} - \frac{\left[r+\gamma \mu_{s',2} -\- \bar{\mu}_{\tau,1}\right]^2_+}{2(\bar{\sigma}_{\tau,1}^2 + \gamma^2 \sigma_{s',2}^2)} \right\} \nonumber \\
 &\; = \frac{1}{\sqrt{2\pi v_1}}\exp\left\{ -\frac{y_1^2}{2v_1} - \frac{\left[y_2 - \alpha_1 y_1 \right]^2_+}{2v_1^{-1}(v_1v_2-v_0^2)} \right\} \quad \text{  where  } \alpha_b \equiv \frac{\sigma_{s,a}^2}{\sigma_{s,a}^2 + \gamma^2\sigma_{s',b}} = \frac{v_0}{v_b} \label{eq:approx_cph}
 \end{align}
 Since the RHS of the equation is a sum of exponential function with the denominator of the inside term is proportional to a negative inverse variance, $ \E_{q \sim \hat{p}_{Q_{s,a}}(\cdot)}[q]$ is approximated to $\bar{\mu}_{\tau,2} = (1-\alpha_2)\mu_{s,a} + \alpha_2 (r+\gamma\mu_{s',a_2})$ if $c_{\tau,1}\Phi_{\tau,1} \ll c_{\tau,2}\Phi_{\tau,2}$ which is identical with the Q-learning update. Therefore, proving Theorem \ref{theorem:convergence_numeric} is equivalent to proving the following statement. If $\mu_{s',2} > \mu_{s',1}$, and $\sigma_{s,a}, \sigma_{s',1}$, and $\sigma_{s',2}$ approach to 0, then $c_{\tau,1}\Phi_{\tau,1}/ c_{\tau,2}\Phi_{\tau,2}$ approaches to 0.
From the Eq.\ref{eq:mean_approx} and Eq.\ref{eq:approx_cph},
\begin{align}
\log\left(\sqrt{\frac{v_1}{v_2}}\frac{c_1\Phi_1}{c_2\Phi_2} \right) = -\frac{y_1^2}{2v_1} - \frac{\left[y_2-\alpha_1 y_1\right]^2_+}{2v_1^{-1}(v_1 v_2 - v_0^2)} +\frac{y_2^2}{2v_2} + \frac{\left[y_1-\alpha_2 y_2\right]^2_+}{2v_2^{-1}(v_1 v_2 - v_0^2)}  \label{eq:rate}
\end{align}

Here, $\left[y_2-\alpha_1 y_1\right]^2_+$ is 0 if $\bar{\mu}_{\tau,1} \geq r+ \gamma \mu_{s',2}$. Likewise, $\left[y_1-\alpha_2 y_2\right]^2_+$ is 0 if $\bar{\mu}_{\tau,2} \geq r+ \gamma \mu_{s',1}$. We consider the following three cases which determine whether the max function terms are 0 or not.
\begin{enumerate}[(i)]
\item For $\mu_{s,a} < r+\gamma \mu_{s',1}$ and $\bar{\mu}_{\tau,2} < r + \gamma \mu_{s',1}$,
\begin{align*}
(RHS)=&\; -\frac{y_1^2}{2v_1}\left(1 + \frac{v_0^2}{v_1v_2 - v_0^2} - \frac{v_1v_2}{v_1v_2 - v_0^2} \right) + \frac{y_2^2}{2v_2}\left(1 + \frac{v_0^2}{v_1v_2 - v_0^2} - \frac{v_1v_2}{v_1v_2 - v_0^2} \right) \\
=&\; \left( -\frac{y_1^2}{2} + \frac{y_2^2}{2} \right) \cdot 0
\end{align*}
Therefore, 
\[
	\frac{c_1\Phi_1}{c_2\Phi_2} = \sqrt{\frac{v_2}{v_1}} \qquad \text{ and } \qquad \mu_{s,a}^{(new)} = \frac{\bar{\mu}_{\tau,1}\sqrt{v_2} + \bar{\mu}_{\tau,2} \sqrt{v_1}}{\sqrt{v_1} + \sqrt{v_2}}
\]
Since $\bar{\mu}_{\tau,1}\geq \mu_{s,a}$ and $\bar{\mu}_{\tau,2}\geq \mu_{s,a}$, the newly updated mean is located somewhere between $\bar{\mu}_{\tau,1}$ and $\bar{\mu}_{\tau,2}$ and always $\mu_{s,a}^{(new)} \geq \mu_{s,a}$. Therefore, if $\mu_{s,a} < r+\gamma \mu_{s',1}$ and $\bar{\mu}_{\tau,2} \leq r + \gamma \mu_{s',1}$, then $\mu_{s,a}^{(new)} > \mu_{s,a}$ until $\bar{\mu}_{\tau,2}$ becomes larger than $r + \gamma \mu_{s',1}$.
\\
\item For $r+\gamma \mu_{s',1} \leq \bar{\mu}_{\tau,1} < r+\gamma \mu_{s',2}$ ($\bar{\mu}_{\tau,2} > r+\gamma\mu_{s',1}$ from this condition),
\begin{align*} 
(RHS) =&\; -\frac{y_1^2}{2v_1} - \frac{\left(y_2-\alpha_1 y_1\right)^2}{2v_1^{-1}(v_1 v_2 - v_0^2)} +\frac{y_2^2}{2v_2} \\
 = &\;- \frac{\left(y_1-\alpha_2 y_2\right)^2}{2v_2^{-1}(v_1 v_2 - v_0^2)} \\
\end{align*}
Therefore, $(RHS)<0$ and
\[
\lim_{\sigma_{s,a}, \sigma_{s',1}, \sigma_{s',2}  \rightarrow 0} \frac{c_1\Phi_1}{c_2\Phi_2} = \lim_{v_0, v_1, v_2 \rightarrow 0} \left[ \sqrt{\frac{v_2}{v_1}} \exp\left\{ - \frac{\left(y_1-\alpha_2 y_2\right)^2}{2v_2^{-1}(v_1 v_2 - v_0^2)}  \right\} \right] = 0
\]

\item For $\mu_{s,a} > r+\mu_{s',2}$ and $\bar{\mu}_{\tau,1} \geq r+ \gamma \mu_{s',2}$ ($\bar{\mu}_{\tau,2} > r+\gamma\mu_{s',1}$ from this condition),
\begin{align*}
(RHS) =&\; -\frac{y_1^2}{2v_1} + \frac{y_2^2}{2v_2}\\
=&\; -\frac{y_1^2}{2v_1} \left(1 - \frac{v_1}{v_2}\frac{y_2^2}{y_1^2} \right)
\end{align*}
If $y_2^2/v_2 < y_1^2/v_1$, then $(RHS) < 0$ with $\sigma_{s,a}, \sigma_{s',1}, \sigma_{s',2}  \rightarrow 0$, and thus  $c_{\tau,1}\Phi_{\tau,1}/ c_{\tau,2}\Phi_{\tau,2}$ approaches to 0 as the previous case. If $y_2^2/v_2 \geq y_1^2/v_1$, 
\[
	\frac{c_1\Phi_1}{c_2\Phi_2} = C\sqrt{\frac{v_2}{v_1}} \qquad \text{for some constant }C
\]
Therefore,
\[
	\mu_{s,a}^{(new)} = \frac{\bar{\mu}_{\tau,1}C + \bar{\mu}_{\tau,2}}{C + 1}
\]
Similar to the first case,  $\mu_{s,a}^{(new)}$ will be located somewhere between $\bar{\mu}_{\tau,1}$ and $\bar{\mu}_{\tau,2}$ and always $\mu_{s,a}^{(new)} < \mu_{s,a}$ until $\bar{\mu}_{\tau,1}$ becomes smaller than or equal to $r+\gamma \mu_{s',2}$.
\end{enumerate}
In conclusion, when the variables satisfy either (i) or (iii), the mean value is contracted to the range corresponding to (ii) which is identical to the Q-learning update. 

For $\sigma_w>0$, $r+\gamma \mu_{s',b'}$ and $\gamma \sigma_{s',b'}$ in the CDF terms are replaced by $\mu^w_{b'}$ and $\sigma^w_{b'}$, respectively as Eq.\ref{eq:likelihood_sto}. $\sigma^w_{b'}$ approaches 0 as $\sigma_{s,a}, \sigma_{s',1}, \sigma_{s',2}  \rightarrow 0$ and therefore, the above proofs are applied. However, the proofs are invalid when $\sigma_{s',b'}/\sigma_w = 0$ since the CDF terms in the likelihood distribution are no longer functions of $q$. As we have shown in the section \ref{app:sto_asymp}, the posterior mean is:
\[
	\mu_{s,a}^{(new)} = \bar{\mu}_{\tau,b^+} = \bar{\sigma}_{\tau,b^+}^2\left( \frac{\mu_{s,a}}{\sigma_{s,a}^2} + \frac{r+\gamma \mu_{s',b^+}}{\gamma^2 \sigma^2_{s',b^+} + \sigma^2_w} \right) 	
\]
where $b^+ = \argmax_{b \in \mathcal{A}} \mu_{s',b}$. Thus, the update rule is still identical to the Q-learning update rule with the following learning rate, $\alpha_{\tau}$:
\[
	\alpha_{\tau} = \frac{\sigma_{s,a}^2}{\sigma_{s,a}^2 + \gamma^2 \sigma_{s',b^+}^2 + \sigma^2_w}
\]
\end{proof}
\subsection{Theorem 2: Convergence of ADFQ}
\textbf{Theorem 2.} \textit{
The ADFQ update on the mean $\mu_{s,a}$ $\forall s\in \mathcal{S}$, $\forall a \in \mathcal{A}$ for $|\mathcal{A}|=2$ is equivalent to the Q-learning update if the variances approach 0 and if all state-action pairs are visited infinitely often. In other words, we have :
\[
	\lim_{k\rightarrow \infty, \{\sigma\} \rightarrow 0}\mu_{s,a;k+1} = \left(1-\alpha_{\tau;k} \right) \mu_{s,a;k} + \alpha_{\tau;k} \left(r+\gamma \max_{b\in \mathcal{A}} \mu_{s',b;k}\right)
   \]
 where $\alpha_{\tau;k}=\sigma_{s,a;k}^2 / \left( \sigma_{s,a;k}^2 + \gamma^2 \sigma_{s',b^+;k}^2 + \sigma_w^2\right)$ and $b^+ = \argmax_{b\in\mathcal{A}} \mu_{s',b}$.}
\begin{proof}
Similar to the proof for the exact update case, we will show that the ratios of the coefficients, $k^*_b/k^*_{b_{max}}$ becomes 0 $\forall b\in \mathcal{A}$, $b\neq  b_{max}$ where $b_{max} = \argmax_b{\mu_{s',b}}$, and $\mu_b^* \rightarrow \bar{\mu}_b$ as $\sigma_{s,a},\sigma_{s',b}$ $\forall b\in \mathcal{A}$ goes to 0. 
When $|\mathcal{A}| = 2$ and $\mu_{s',2} > \mu_{s',1}$, 
\begin{align*}
\frac{k_1^*}{k_2^*} = \frac{\sigma_1^*}{\sigma_2^*}\frac{\sigma_{s',2}}{\sigma_{s',1}} \exp\left\{ -\frac{y_1^2}{2v_1} + \frac{y_2^2}{2v_2} - \frac{(\mu_1^*-\bar{\mu}_{\tau,1})^2}{2\bar{\sigma}^2_{\tau,1}} \right. &\;  +\frac{(\mu_2^*-\bar{\mu}_{\tau,2})^2}{2\bar{\sigma}^2_{\tau,2}}  \\
&\;  \left. - \frac{[r+\gamma \mu_{s',2} - \mu_1^*]_+^2}{2\gamma^2\sigma_{s',2}^2} + \frac{[r+\gamma \mu_{s',1} - \mu_2^*]_+^2}{2\gamma^2\sigma_{s',1}^2}\right\}
\end{align*}
According to the definition of $\mu_b^*$,
\[
	\mu_1^*-\bar{\mu}_{\tau,1} = \frac{\bar{\sigma}_{\tau,1}}{\gamma^2\sigma^2_{s',2}} [r + \gamma \mu_{s,'2} - \mu_1^*]_+
\]
and $\mu_b^* \geq \bar{\mu}_{\tau,b}$.
Therefore,
\begin{align*}
\log\left(\frac{k_1^*\sigma_{s',1}\sigma_2^*}{k_2^*\sigma_{s',2}\sigma_1^*} \right) = -\frac{y_1^2}{2v_1} + \frac{y_2^2}{2v_2}  - &\; \frac{[r+\gamma \mu_{s',2} - \mu_1^*]_+^2}{2\gamma^2\sigma_{s',2}^2}\left(1 + \frac{\bar{\sigma}_{\tau,1}^2}{\gamma^2\sigma_{s',2}^2} \right)\\ &\;\qquad + \frac{[r+\gamma \mu_{s',1} - \mu_2^*]_+^2}{2\gamma^2\sigma_{s',1}^2}\left(1 + \frac{\bar{\sigma}_{\tau,2}^2}{\gamma^2\sigma_{s',1}^2} \right)
\end{align*}
When $\mu_b^* < r+\gamma \mu_{s',b'}$
\begin{equation}
	\mu_b^* = \left(\frac{1}{\bar{\sigma}_{\tau,b}} + \frac{1}{\gamma^2 \sigma_{s',b'}^2} \right)^{-1} \left(\frac{\bar{\mu}_{\tau,b}}{\bar{\sigma}_{\tau,b}} + \frac{r+\gamma \mu_{s',b'}}{\gamma^2 \sigma_{s',b'}^2} \right) \label{eq:mu_star4two}
\end{equation}
When $\mu_b^* \geq r+\gamma \mu_{s',b'}$, $\mu_b^* = \bar{\mu}_{\tau,b}$. 

For $\mu_1^* < r+\gamma \mu_{s',2}$ and $\mu_2^* < r+\gamma \mu_{s',1}$, it is also, $\mu_{s,a} \leq \bar{\mu}_{\tau,1} \leq \bar{\mu}_{\tau,2} < r+\gamma \mu_{s',1} < r+ \gamma \mu_{s',2}$. Then, using Eq.\ref{eq:mu_star4two}, we have
\begin{align*}
\log\left(\frac{k_1^*\sigma_{s',1}\sigma_2^*}{k_2^*\sigma_{s',2}\sigma_1^*} \right) = -\frac{y_1^2}{2v_1} + - \frac{\left(y_2-\alpha_1 y_1\right)^2}{2v_1^{-1}(v_1 v_2 - v_0^2)} +\frac{y_2^2}{2v_2} + \frac{\left(y_1-\alpha_2 y_2\right)^2}{2v_2^{-1}(v_1 v_2 - v_0^2)} 
\end{align*}
which is same with (i) of the proof of Theorem 1. The new mean will be weighted sum of $\mu_1^*$, $\mu_2^*$. Since $\mu_{s,a}$ is smaller than both $\bar{\mu}_{\tau,1}$ and $\bar{\mu}_{\tau,2}$, $\mu_{s,a}^{(new)} > \mu_{s,a}$ until $r+\gamma_{s',1} < \bar{\mu}_{\tau,2}$. For the other cases, the same directions in the proof of Theorem 1 are applied.

We can apply the same proof procedures to the case of $\sigma_w>0$ using $\gamma^2 \sigma_{s',b}^2 + \sigma_w^2$ instead of $\gamma^2 \sigma_{s',b}^2$ in $\bar{\mu}_{\tau,b}$, $\bar{\sigma}_{\tau,b}$, and $c_{\tau,b}$. Therefore, the mean update rule converges to the Q-learning update and the corresponding learning rate is:
\[
	\alpha_{\tau} = \frac{\sigma_{s,a}^2}{\sigma_{s,a}^2 + \gamma^2 \sigma_{s',b^+}^2 + \sigma^2_w}
\]
\end{proof}
\section{Experimental Details} \label{app:exp_details}
\subsection{Neural Network Architecture and Details}
In the all domains, we used the default settings of the OpenAI baselines \cite{baselines} for DQN and Double DQN, and made minimal changes for ADFQ. We used ReLU nonlinearities and the Adam optimizer with mini-batches size of 32. 

\subsection{Initialization} In ADFQ, Xavier initialization was used for all weight variables, and all bias variables were initialized to zero except for the final hidden layer. The weights of the final hidden layer were initialized with $0.0$ and its bias variables were initialized with two constant values which correspond to $\mu_0$ and $-\log(\sigma_0)$ where $\mu_0$ is an initial mean and $\sigma_0^2$ is an initial variance (e.g. an initial bias vector of the final layer is $\vec{b} = [\mu_0, \cdots, \mu_0,-\log(\sigma_0), \cdots, -\log(\sigma_0)]^T$). We set $\sigma_0 =50.0$ for the Atari games. 

\subsection{Network Architecture} We used a network with three convolution layers followed by a 256 neuron linear layer. The first convolution layer contains 32 filters of size 8 with stride 4. The second convolution layer contains 64 filters of size 4 with stride 2. The final convolution layer contains 64 filters of size 3 with stride 1. This network architecture is the default setting in OpenAI baselines for the breakout example.

\section{Additional Experiments in Discrete MDPs} \label{app:add_experiments}
In this section, we evaluate ADFQ in MDPs with finite state and action spaces without using a function approximator. We show the convergence of ADFQ as well as its performance, and compare with Q-learning and KTD-Q.
\subsection{Algorithms}
ADFQ is evaluated with two action policies: \textit{Thompsing Sampling (TS)} \cite{thompson} selects $a_t = \argmax_a q_{s_t,a}$ where $q_{s_t,a} \sim p_{Q_{s_t,a}}(\cdot|\theta_t)$, and $\epsilon$-\textit{greedy} selects a random action with $\epsilon$ probability and selects the action with the highest mean otherwise. In implementation, we fixed the initial variance to 100.0 and the variances are bounded by $10^{-10}$ since their values dramatically drop and eventually exceed the precision range of computers. 

For comparison, we test Q-learning with $\epsilon$-greedy and Boltzmann action policies. The learning rate decreases as the number of visits to a state-action pair increases $\alpha_t = \alpha_0(n_0+1)/(n_0+t)$, $\alpha_0=0.5$ \citep{Lagoudakis}. Additionally, KTD-Q with $\epsilon$-greedy and its active learning scheme are also examined. KTD-Q is an extension of Kalman Temporal Difference (KTD) \citep{Geist} and one of the recent influential algorithms for Bayesian off-policy TD learning. KTD approximates the value function using the Kalman filtering scheme, and KTD-Q handles the non-linearity in the Bellman optimality equation by applying the Unscented Transform. The same hyperparameter values as the ones in the original paper are used if presented. All other hyperparameters are selected through cross-validation.
\subsection{Domains}
\begin{figure}[b!]
    \centering
    \includegraphics[height=4cm]{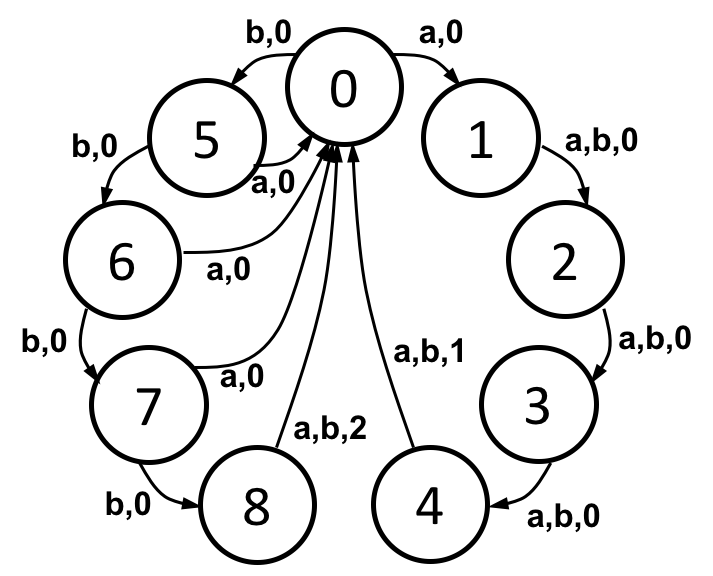}
    \qquad \qquad
    \includegraphics[height=4cm]{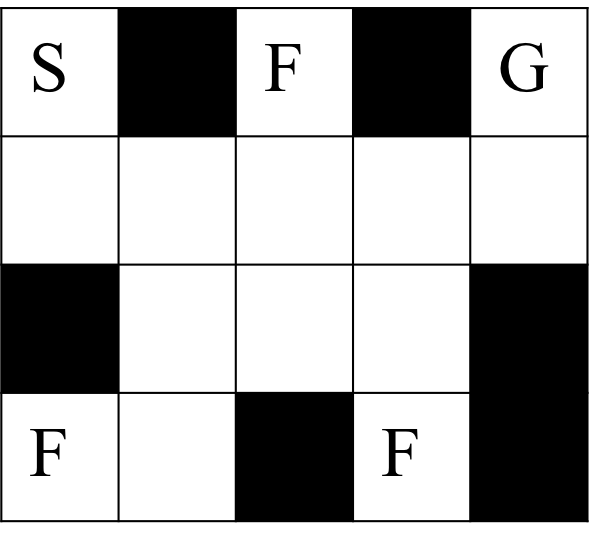}
    \caption{Loop and Maze domain diagrams}
    \label{fig:domains}
\end{figure}

We test the algorithms in Loop and Maze ($\gamma=0.95$, Figure \ref{fig:domains}) presented in \citep{Dearden1} with and without stochasticity in the domains for finite learning steps ($T_{H, loop}=10000$, $T_{H, maze}=30000$). The Loop domain consists of 9 states and 2 actions (a,b). There are +1 reward at state 4 and +2 reward at state 8. For a stochastic case, a learning agent performs the other action with a probability 0.1.
In Maze, the agent's goal is to collect the flags "F" and escape the maze through the goal position "G" starting from "S". It receives a reward equivalent to the number of flags it has collected at "G". The agent remains at the current state if it performs an action toward a wall (black block). For a stochastic case, the agent slips with a probability 0.1 and moves to the right perpendicular direction.
\subsection{Results}
\begin{figure}[tb!]
\centering
  \includegraphics[width=0.45\textwidth]{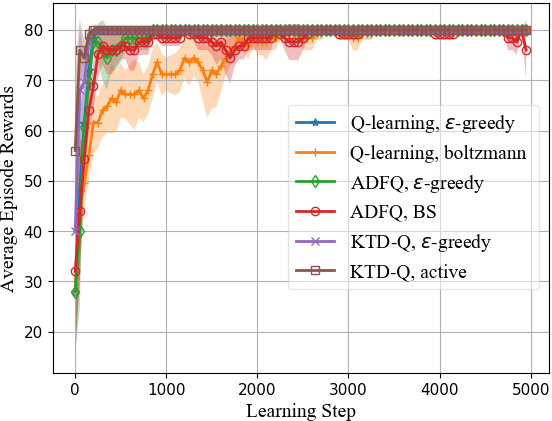}
\qquad
\includegraphics[width=0.45\textwidth]{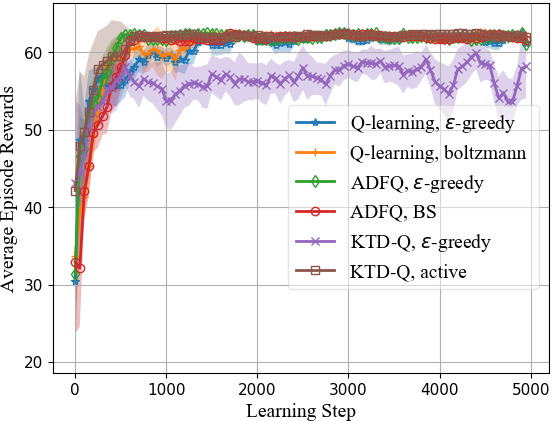}\\
  \includegraphics[width =0.45\textwidth]{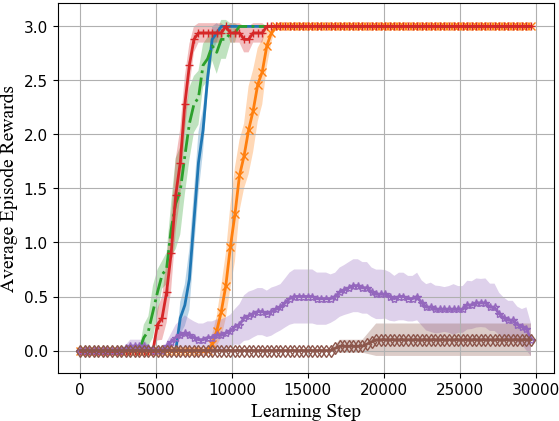}
  \qquad
  \includegraphics[width =0.45\textwidth]{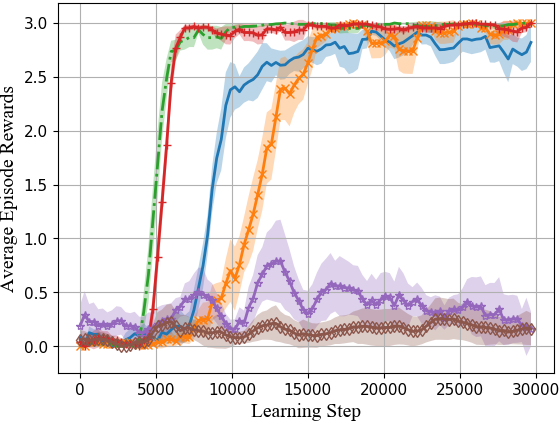}
  \caption{\footnotesize Semi-greedy evaluation during learning smoothed by a moving average with window 4. Left: deterministic, Right: stochastic, Top: Loop, Bottom: Maze.}
  \label{fig:discrete_eval}
\end{figure}

We first examined the convergence to the optimal Q-values using randomly generated fixed trajectories $<s_0,a_0,r_0,s_1,\cdots>$ for all algorithms in order to evaluate only the update part of each algorithm. During learning, we computed the root mean square error (RMSE) between the estimated Q-values (or means) and the true optimal Q-values, and the results were averaged over 10 trials. The true optimal Q-values were obtained using the policy iteration method . 
In addition, we evaluated the performance of each algorithm with different action policies during learning. At every $T_H/100$ steps, the current policy was greedily evaluated where the maximum number of steps was bounded by 1.5 times of the optimal path length or it was terminated when the goal was reached. The entire experiment was repeated 10 times for each domain and the averaged results were plotted in Fig.\ref{fig:discrete_eval}. 

\begin{figure}[tb!]
\centering
  \includegraphics[width =0.45\textwidth]{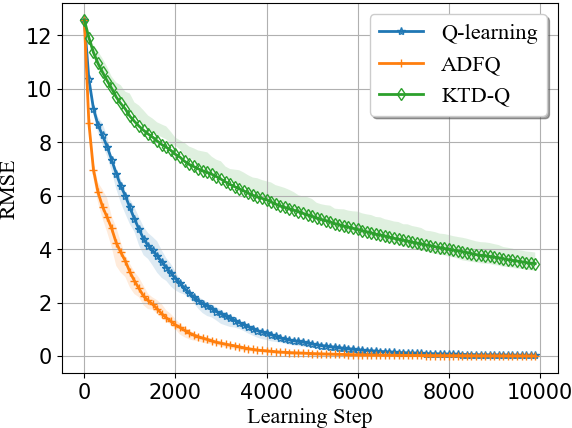}\qquad
\includegraphics[width =0.45\textwidth]{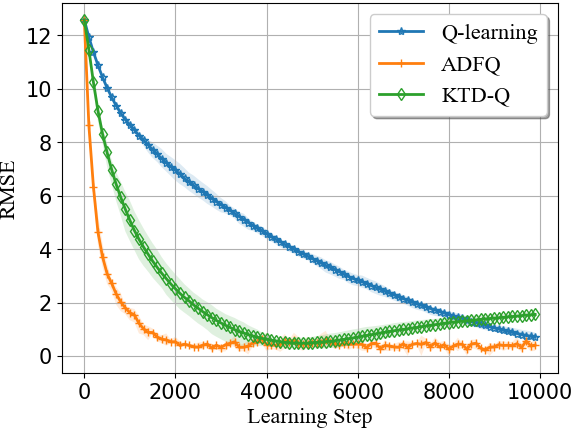}\\
\includegraphics[width =0.45\textwidth]{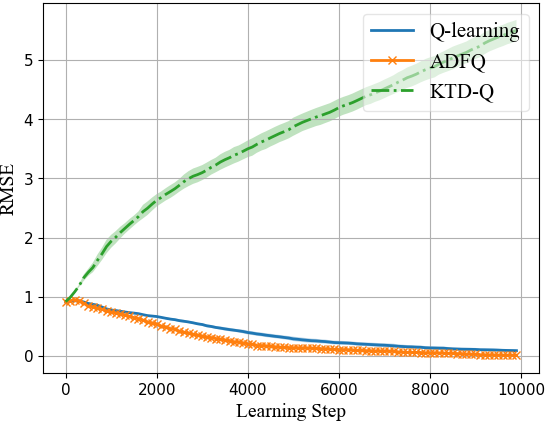}
  \qquad
  \includegraphics[width =0.45\textwidth]{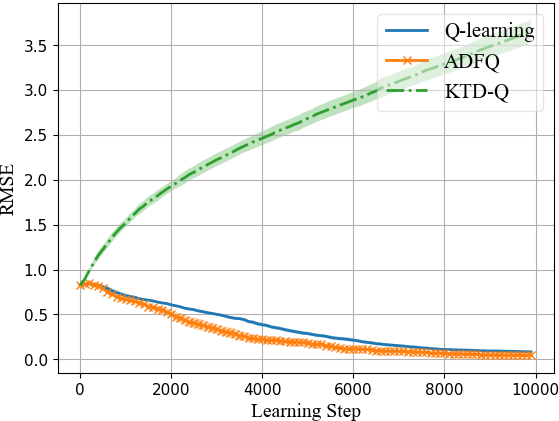}
\caption{\footnotesize Root Mean Square Error (RMSE) of $Q$ or $\mu$ from the optimal Q-values. Left: deterministic, Right: stochastic, Top: Loop, Bottom: Maze.}
\label{fig:discrete_rmse}
\end{figure}

As shown in Fig.\ref{fig:discrete_rmse}, ADFQ converged to the optimal Q-values quicker than all other algorithms including Q-learning. Moreover, ADFQ with $\epsilon$-greedy and ADFQ with \textit{TS} showed similar results and converged to the optimal performance faster than the comparing algorithms in all cases. Q-learning with $\epsilon$-greedy learned an optimal policy almost as fast as ADFQ in the deterministic cases, but the performance of ADFQ was improved dramatically in the stochastic cases. 
KTD-Q approached to the optimal values in the deterministic Loop domain, but diverged in others since its derivative-free approximation nature does not scale well with the number of parameters. It is also proposed under a deterministic environment assumption. The author proposed XKTD-V and XKTD-SARSA which are extended versions of KTD-V and KTD-SARSA, respectively, for a stochastic environment. Yet, KTD-Q was not able to be extended to XKTD-Q (see the section 4.3.2 in \citep{Geist} for details). Despite the convergence issue, the KTD-Q with active learning scheme worked better than Q-learning and converged to an optimal policy in Loop. These results imply that KTD-Q does not scale with the number of parameters even though it works well in smaller domains, and its convergence to the optimal Q-values is not guaranteed in stochastic domains. 


\end{document}